\documentclass[final]{cvpr}

\usepackage{times}
\usepackage{epsfig}
\usepackage{graphicx}
\usepackage{amsmath}
\usepackage{amssymb}
\usepackage{bm}
\usepackage{xcolor}
\usepackage{multirow}

\usepackage[pagebackref=true,breaklinks=true,colorlinks,bookmarks=false]{hyperref}

\begin{document}

%%%%%%%%% TITLE
\title{Diverse Semantic Image Synthesis via Probability Distribution Modeling}

\author{Zhentao Tan$^{1}$, Menglei Chai$^{2}$, Dongdong Chen$^{3}$, Jing Liao$^{4}$, \\ Qi Chu$^{1}$, Bin Liu$^{1}$, Gang Hua$^{5}$, Nenghai Yu$^{1}$\\
$^{1}$University of Science and Technology of China \quad 
$^{2}$Snap Inc.\quad$^3$Microsoft Cloud AI\\
$^4$City University of Hong Kong
\quad$^5$Wormpex AI Research LLC\\
{\tt\small\{tzt@mail., qchu@, flowice@, ynh@\}ustc.edu.cn }\\
{\tt\small mchai@snap.com}, 
{\tt\small \{cddlyf,ganghua\}@gmail.com},
{\tt\small jingliao@cityu.edu.hk}
}

\maketitle

%%%%%%%%% ABSTRACT
\begin{abstract}
Semantic image synthesis, translating semantic layouts to photo-realistic images, is a one-to-many mapping problem.
Though impressive progress has been recently made, diverse semantic synthesis that can efficiently produce semantic-level multimodal results, still remains a challenge.
In this paper, we propose a novel diverse semantic image synthesis framework from the perspective of semantic class distributions, which naturally supports diverse generation at semantic or even instance level. We achieve this by modeling class-level conditional modulation parameters as continuous probability distributions instead of discrete values, and sampling per-instance modulation parameters through instance-adaptive stochastic sampling that is consistent across the network. Moreover, we propose prior noise remapping, through linear perturbation parameters encoded from paired references, to facilitate supervised training and exemplar-based instance style control at test time. Extensive experiments on multiple datasets show that our method can achieve superior diversity and comparable quality compared to state-of-the-art methods. Code will be available at \url{https://github.com/tzt101/INADE.git}
\end{abstract}

%%%%%%%%% BODY TEXT
\section{Introduction}
\begin{figure}[tp]
  \centering
  \includegraphics[width=\linewidth]{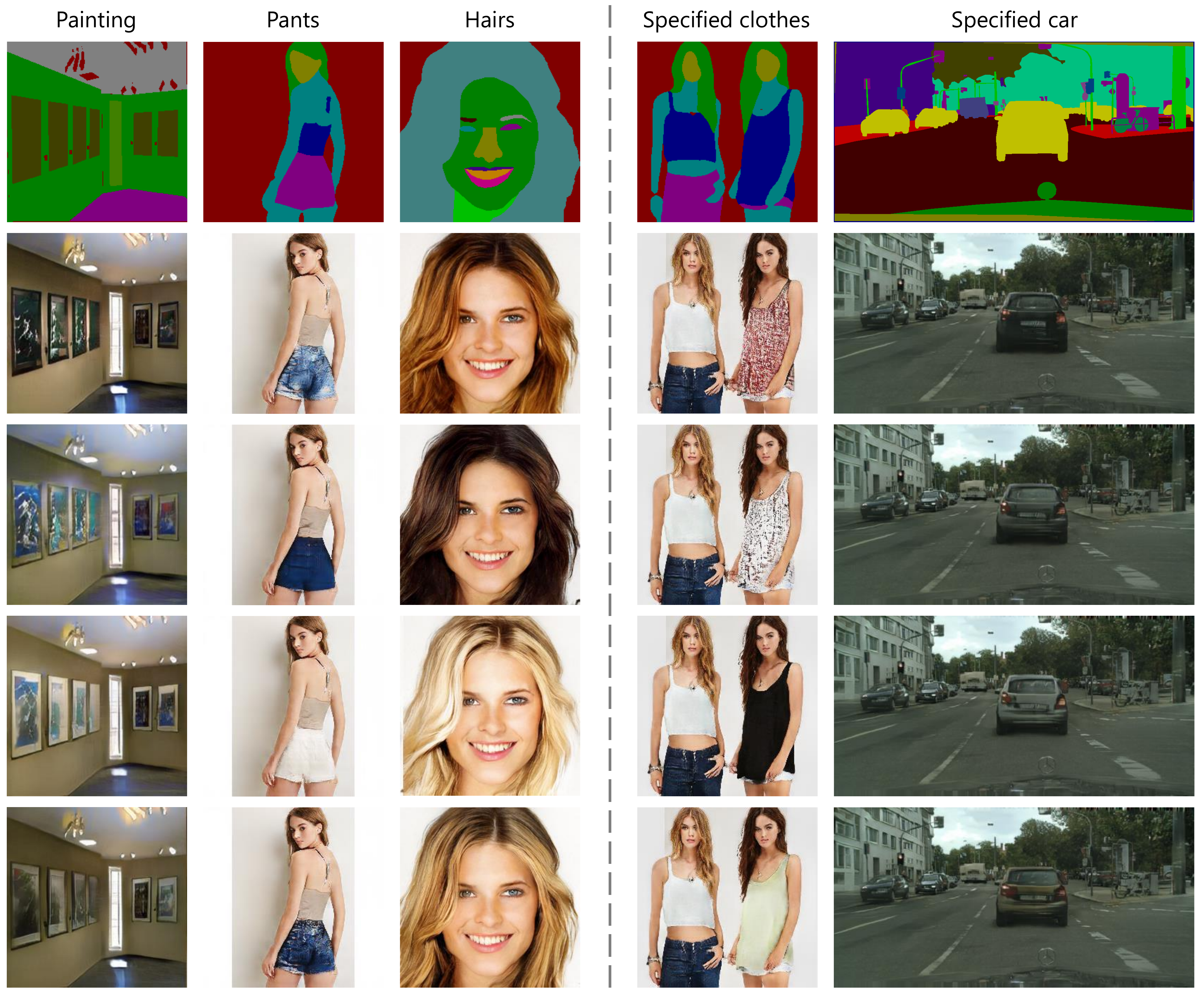}
  \caption{Semantic-level (left three columns) and Instance-level (right two columns) multimodal images generated by the proposed method. Text of each column indicates which semantic class or instance will be changed in the following results.}
  \label{fig:teaser}
\end{figure}

Image synthesis has recently seen impressive progress, particularly with the help of generative adversarial networks (GANs)~\cite{goodfellow2014generative}. Besides stochastic approaches that generate high-quality images from random latent variables~\cite{karras2019style,karras2020analyzing}, conditional image synthesis is attracting equal or even more attention due to the practical advantages of its controllability. The conditional input, to guide the synthesis, can be of various forms, including RGB images, edge/gradient maps, semantic labels, etc. In this work, semantic image synthesis is one particular task that aims to generate a photo-realistic image from a semantic label mask. In particular, we further explore its diversity and controllability without loss of generation quality. Some samples are shown in Figure~\ref{fig:teaser}.

Previous works~\cite{isola2017image,wang2018high} propose solutions within the general image-to-image translation framework, which directly feeds the semantic mask into the encoder-decoder network. For higher quality, some recent methods~\cite{park2019semantic,zhu2020sean,tan2020michigan} adopt spatially-varying conditional normalization to avoid the loss of semantic information due to conventional normalization layers~\cite{ulyanov2016instance}. Although proven successful in synthesizing certain types of content, these methods lack controllability over the generation diversity, which is particularly important for such a one-to-many problem. Some methods~\cite{zhu2017toward,yang2018diversity} attempt to yield multimodal results by incorporating variational auto-encoder (VAE) or introducing noises. However, these methods only support global image-level diversity.
To obtain finer-grained controllability, a recent work~\cite{zhu2020semantically} proposes to use group convolution for different semantics to achieve semantic-level diversity. However, it is computationally expensive and difficult to be extended to support diversity at the instance level.

In this paper, we attempt to achieve controllable diversity in semantic image synthesis from the perspective of semantic probability distributions. The intuition is to treat each semantic class as one distribution, so that each instance of this class could be drawn from this distribution as a discrete sample. Following this idea, we propose a novel semantic image synthesis framework, which is naturally capable of producing diverse results at semantic or even instance level.

Specifically, our method contains three key ingredients. Firstly, we propose \textit{variational modulation models} (\S~\ref{sec:model}) that extend discrete modulation parameters to class-wise continuous probability distributions, which embed diverse styles of each semantic category in a class-adaptive manner. Secondly, based on the variational models built per normalization layer, we further develop an \textit{instance-adaptive sampling} method (\S~\ref{sec:sampling}) that achieves instance-level diversity by stochastically sampling modulation parameters from the variational models. We harmonize the sampling across the network via consistent randomness and a learnable transformation function for each normalization layer. Finally, to more efficiently embed the instance diversity to the modulation models, we propose \textit{prior noise remapping} (\S~\ref{sec:noise}) that transforms the noise samples with perturbation parameters encoded from arbitrary references. We adopt this step to facilitate supervised training and enable test-time reference-based style guidance. Inspired by ~\cite{park2019semantic,tan2020rethinking,tan2020semantic}, the proposed method is called INADE (\textbf{IN}stance-\textbf{A}daptive \textbf{DE}normalization).

To evaluate the proposed method, we conduct extensive experiments on multiple datasets, including Cityscapes~\cite{cordts2016cityscapes}, ADE20K~\cite{zhou2017scene}, CelebAMask-HQ~\cite{lee2020maskgan,karras2017progressive,liu2015deep}, and DeepFashion~\cite{liu2016deepfashion}. Both quantitative and qualitative results show that our method significantly outperforms state-of-the-art methods by achieving much better instance-level diversity while keeping comparable generation quality. 

%------------------------------------------------------------------------
\section{Related Work}
\subsection{Conditional Image Synthesis}
Conditional image synthesis aims at generating photo-realistic images conditioned on different types of input. We are interested in a special form of it, called semantic image synthesis, which takes segmentation layouts as input. Many impressive works have been proposed for this task. The most representative work, Pix2Pix~\cite{isola2017image} adopts an encoder-decoder generator for unified image-to-image translation. Pix2pixHD~\cite{wang2018high} improves Pix2Pix by proposing coarse-to-fine generator and discriminators. 
Subsequent methods~\cite{qi2018semi,liu2019learning,tang2020dual,zhang2020cross,tan2020semantic,zhu2020sean} further explore how to synthesize high quality images from semantic masks and achieve significant improvements.
Besides using class-level semantic masks, some works also consider instance-level information for image synthesis, since the semantic mask itself does not provide sufficient information to synthesize instances especially in complex environments with multiple of them interacting with each other. 
Some works~\cite{wang2018high,park2019semantic,tan2020semantic} extract boundary information from the instance map and concatenate it with the semantic mask. While recent work~\cite{dundar2020panoptic} proposes to use the instance map to guide convolution and upsampling layers for better exploiting both semantic and instance information. 
Different from these methods, we are interested in taking full advantage of information from instance maps to achieve instance-level diversity control.

\subsection{Diversity in Image Synthesis}
Diversity is a core target for image synthesis, which aims to generate multiple possible outputs from a single input image. Early conditional image synthesis networks either trained with paired data, like Pix2Pix~\cite{isola2017image} and Pix2pixHD~\cite{wang2018high}, or with unpaired data, like CycleGAN~\cite{zhu2017unpaired}, DiscoGAN~\cite{kim2017learning} and UNIT \cite{liu2017unsupervised}, are single-modal. They produce one single output conditioned solely on an input image. Later, some multimodal unpaired image synthesis networks~\cite{huang2018multimodal,lee2018diverse,almahairi2018augmented} are proposed. 
However, constrained by the reconstruction loss, the semantic image synthesis task trained with paired data is more difficult to support diversity. 
Simply concatenating a random noise vector to the input segmentation mask is usually not effective, because the generator often ignores the additional noise vectors and mode collapse may occur easily. To tackle this problem, BicycleGAN~\cite{zhu2017toward} enforces the bijection mapping between the noise vector and target domain. DSCGAN~\cite{yang2018diversity} proposes a simple regularization which can be easily integrated into most conditional GAN objectives. 
More recently, a variational autoencoder architecture is used to handle multimodal synthesis by~\cite{park2019semantic,tan2020semantic,liu2019learning}.
However, these multimodal image synthesis networks only support diversity at the global level. 
To further control the diversity at the semantic level, the method proposed by~\cite{gu2019mask} builds several auto-encoders for each face component to extract different component representations. 
GroupDNet~\cite{zhu2020semantically} unifies the generation process in only one model, but still requires high computing resources, and the use of group convolution layer makes it difficult to extend to the instance level.
In contrast, we propose a novel instance-aware conditional normalization framework that allows diverse instance-level generation with less overhead.

\begin{figure*}[t]
  \centering
  \includegraphics[width=0.92\linewidth]{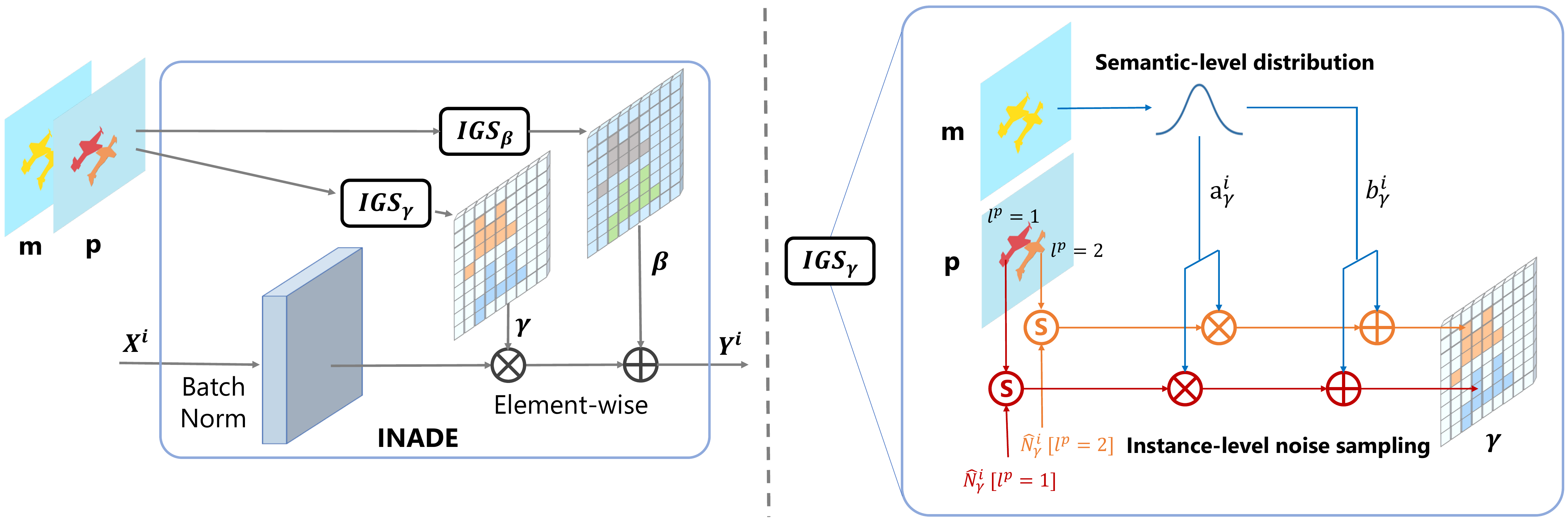}
  \caption{The illustration diagram of the proposed \textbf{IN}stance-\textbf{A}daptive \textbf{DE}normalization (INADE). It combines semantic-level distribution modeling and instance-level noise sampling. IGS denotes the \textbf{I}nstance \textbf{G}uided \textbf{S}ampling which is similar to the guided sampling in~\cite{tan2020semantic}.}
  \label{fig:INADE}
\end{figure*}
%------------------------------------------------------------------------
\section{Method}
We are interested in the task of semantic image synthesis, which is defined as to map a semantic mask $\bm{m}\in\mathbb{L}_m^{H\times W}$ to a photo-realistic image $\bm{o}\in\mathbb{R}^{3\times H\times W}$. Here, $\bm{m}$ is a class-level label map with each pixel representing an integer index to a pre-defined set of semantic categories $\mathbb{L}_m=\{1,2,\ldots,L^m\}$. Each pair of input $\bm{m}$ and output $\bm{o}$ is spatially-aligned and of the same dimension $H\times W$, so that the synthesized content in $\bm{o}$ should comply with the corresponding semantic labels in $\bm{m}$.

In addition to this basic formulation~\cite{wang2018high,park2019semantic,tan2020semantic}, instance-aware semantic image synthesis~\cite{dundar2020panoptic} adopts the instance map $\bm{p}\in\mathbb{L}_p^{H\times W}$ as an extra input, which differentiates different object instances sharing a same semantic label by denoting each individual instance with a unique index from the instance label set $\mathbb{L}_p=\{1,2,\ldots,L^p\}$ in the image. By enforcing an identical semantic label within each instance, pixels belonging to a same instance label $l^p$ in $\bm{p}$ should always have a same semantic label $l^m$ in $\bm{m}$. We represent instance to semantic label mapping as a function $l^m=\mathcal{G}(l^p)$.

Overall, image synthesis with instance information can be basically formulated as a function $\mathcal{T}(\bm{m},\bm{p}):(\mathbb{L}_m,\mathbb{L}_p)\to\mathbb{R}^3$. And feed-forward image translation neural networks, trained in a supervised manner, can be used to model this function. In the following, we introduce the proposed method with both the inputs of semantic and instance maps. When there is no instance label, $\bm{p}$ degenerates into $\bm{m}$. And the diversity of the synthesized images changes from the instance level to the semantic level.

 \begin{figure*}[t]
  \centering
  \includegraphics[width=\linewidth]{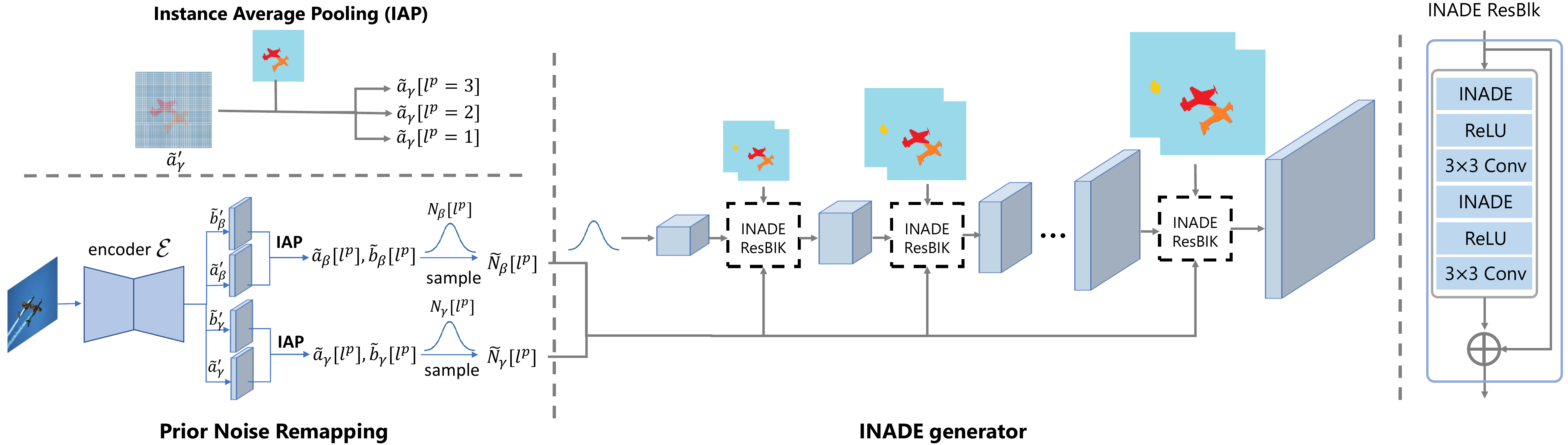}
  \caption{The overall framework of the proposed INADE generator, which consists of a remapping encoder $\mathcal{E}$ and INADE generator. $\mathcal{E}$ is used to transform the noise sample based on arbitrary references (\S~\ref{sec:noise}), while the generator consists of several INADE ResBlks.}
  \label{fig:generator}
\end{figure*}

\subsection{Conditional Normalization}
\label{sec:normalization}
Our solution to semantic image synthesis is based on a novel instance-level conditional normalization method. Before introducing our method, here we give a brief overview of the general framework of conditional normalization first.

Let $\bm{X}^i\in\mathbb{R}^{C^i\times H^i\times W^i}$ be the activation tensor to the $i$-th normalization layer, where $C^i, H^i, W^i$ are the channel depth, height, and width, respectively. In the channel-wise normalization framework similar to~\cite{ioffe2015batch}, we can generally formulate the normalization operations as two steps: In the normalization step, $\bm{X}^i$ is normalized to $\bm{\hat{X}}^i$ by channel-wise mean and standard deviation $\{\bm{\mu}^i,\bm{\sigma}^i\}\in\mathbb{R}^{C^i}$ in the mini-batch containing $\bm{X}^i$. Then, the modulation step scales and translates $\bm{\hat{X}}^i$ with learned modulation parameters $\{\bm{\gamma}^i,\bm{\beta}^i\}\in\mathbb{R}^{C^i\times H^i\times W^i}$, which are not necessarily channel-wise constant. Let $\bm{Y}^i$ be the output, for each element ($k\in C^i, x\in H^i, y\in W^i$) in the tensor, we have:
\begin{equation}
\begin{split}
  \bm{\hat{X}}^i_{k,x,y}=(\bm{X}^i_{k,x,y}-\bm{\mu}^i_k)/\bm{\sigma}^i_k, \\
  \bm{Y}^i_{k,x,y}=\bm{\gamma}^i_{k,x,y}\bm{\hat{X}}^i_{k,x,y}+\bm{\beta}^i_{k,x,y}.
 \end{split}
\end{equation}

For conditional normalization~\cite{dumoulin2016learned,huang2017arbitrary}, the modulation parameters $\bm{\gamma}^i$ and $\bm{\beta}^i$ are learned with extra conditions. Specifically, for semantic image synthesis, the modulation is usually conditioned on the semantic mask $\bm{m}$~\cite{park2019semantic,tan2020semantic}.

\subsection{Variational Modulation Model}
\label{sec:model}
Conditional normalization (\S~\ref{sec:normalization}), either spatially-adaptive~\cite{park2019semantic} or class-adaptive~\cite{tan2020semantic}, has been proven helpful for semantic image synthesis. The semantic-conditioned modulation is able to largely prevent the "wash-away" effect of semantic information caused by repetitive normalizations. However, challenges still exist to achieve promising generation results with semantic-level or even instance-level diversity, given that normalization is solely conditioned on the semantic map and only global randomness is used to diversify the image styles~\cite{park2019semantic}. Semantic-level diversity is realized by~\cite{zhu2020semantically} through group convolution, but using this convolution cuts off the possibility of its extension to instance-level diversity through instance map.
Recent efforts on instance-aware synthesis~\cite{wang2018high,dundar2020panoptic} are majorly focused on better object boundaries, but not the diversity and realism of each individual instance. Due to the lack of proper instance conditioning, existing methods tend to converge instances with the same semantic label into a similar style, which significantly harms the diversity of generation. 

The key to instance-level diversity is a proper combination of uniform semantic-level distributions that deterministically decide the general features of a particular semantic label, and instance-level randomness that introduces allowed diversity covered by the semantic distribution models. Therefore, we model the modulation parameters as parametric probability distributions for each semantic label $l^m\in\mathbb{L}^m$, instead of discrete values. With such a, namely, variational modulation model, given an instance $l^p\in\mathbb{L}^p$, instance-level diversity is achievable via sampling modulation parameters from the probability distributions of the corresponding semantic label $\mathcal{G}(l^p)$. For the sake of simplicity and efficiency, following~\cite{tan2020semantic}, we make the modulation parameters spatially-invariant and only depend on the local instance labels.

Specifically, for each semantic category $l^m$, its channel-wise modulation parameters are modeled as learnable probability distributions, which are built for each normalization layer in the network respectively. Formally, for the $i$-th layer with channel depth $C^i$, we have $\{\bm{a}^i_{\gamma},\bm{b}^i_{\gamma},\bm{a}^i_{\beta},\bm{b}^i_{\beta}\}\in\mathbb{R}^{L^m\times C^i}$ as the distribution transformation parameters of $\gamma$ and $\beta$, respectively. All of them are treated as learnable parameters that are jointly trained with the network. Given stochastic noise matrices $\{\bm{N}^i_\gamma,\bm{N}^i_\beta\}\in\mathbb{R}^{L^p\times C^i}$ from the same distribution for sampling, the corresponding modulation parameters of one instance label $l^p$ in $\bm{p}$ are:
\begin{equation}
\begin{split}
  \bm{\gamma}^i[l^p]&=\bm{a}^i_\gamma[\mathcal{G}(l^p)]\otimes\bm{N}^i_\gamma[l^p]+\bm{b}^i_\gamma[\mathcal{G}(l^p)],\\
  \bm{\beta}^i[l^p]&=\bm{a}^i_\beta[\mathcal{G}(l^p)]\otimes\bm{N}^i_\beta[l^p]+\bm{b}^i_\beta[\mathcal{G}(l^p)],
\end{split}
\label{eq:sample}
\end{equation}
where $\otimes$ represents element-wise multiplication, and $[\cdot]$ accesses the vector from a matrix in the row-major order. 

\subsection{Instance-Adaptive Modulation Sampling}
\label{sec:sampling}
Our multimodal synthesis method follows the basic form of conditional modulation (\S~\ref{sec:normalization}), but further extends the conditional inputs to include not just the segmentation mask $\bm{m}$, but also the instance map $\bm{p}$ and random noises to initiate sampling, as shown in Figure~\ref{fig:INADE}. Utilizing our variational modulation models (\S~\ref{sec:model}), we are able to generate diverse modulation parameters obeying the same set of probability distributions. However, considering that the generation network contains multiple conditional normalization layers, a unified sampling solution is still essential to harmonize all these layers. A straight-forward approach, independent stochastic sampling for each normalization layer, could potentially introduce inconsistency and cause the diversity to be severely neutralized. Therefore, in this paper, we propose an instance-adaptive modulation sampling method that achieves consistent instance sampling across multiple normalization layers with unequal channel depths.

To initialize, for each layer $i$, we resize and convert each input pair of semantic mask $\bm{m}$ and instance map $\bm{p}$ into the one-hot format as $\bm{M}^i\in\mathbb{B}^{L^m\times H^i\times W^i}$ and $\bm{P}^i\in\mathbb{B}^{L^p\times H^i\times W^i}$, respectively, which will then be used as the conditional inputs to that layer, as shown in Figure~\ref{fig:generator}. Here, $\mathbb{B}$ represents the Boolean domain, and $L^m,L^p$ are the aforementioned total numbers of semantic/instance labels.

For the sake of simplicity, since scale $\bm{\gamma}^i$ and shift $\bm{\beta}^i$ are generated similarly and independently, without loss of generality, here we take scale $\bm{\gamma}^i$ as the example. The sampling contains the following steps.

First of all, random samples $\bm{N}_\gamma\in\mathbb{R}^{L^p\times C^0}$ are independently sampled from the standard normal distribution: $\bm{N}_\gamma\sim\mathcal{N}(0,1)$. We use the same set of random noise samples $\bm{N}_\gamma$ for all instances of normalization layers in the network, which helps enforce consistent instance styles throughout the network. Here $C^0$ is a hyper-parameter that defines the number of the initial sampling channels.

To sample the modulation parameters for each normalization layer $i$, we translate the initial samples $\bm{N}_\gamma$ to $\hat{\bm{N}}^i_\gamma$ with a learnable linear transformation mapping $\mathcal{F}^i_\gamma:\mathbb{R}^{L^p\times C^0}\to\mathbb{R}^{L^p\times C^i}$:
\begin{equation}
    \hat{\bm{N}}^i_\gamma = \mathcal{F}^i_\gamma(\bm{N}_\gamma),
\end{equation}
where $C^i$ is exactly the channel depth of $i$-th activations $\bm{X}^i$, so that the output $\hat{\bm{N}}^i_\gamma\in\mathbb{R}^{L^p\times C^i}$ assigns a transformed sample for each instance per each channel, in a spatially-invariant manner. Thus, the same source of randomness helps achieve style consistency, while the learnable transformations enforce compatible target dimensions and preserve certain ability to adapt the samples for each layer.

Finally, given the distribution transformation parameters $\bm{a}^i_{\gamma},\bm{b}^i_{\gamma}$ and transformed noise samples $\hat{\bm{N}}^i_\gamma$, the scale parameters $\bm{\gamma}^i$ are calculated with Equation~\ref{eq:sample}. And similarly for the shift parameters $\bm{\beta}^i$.

\subsection{Prior Noise Remapping}
\label{sec:noise}
While our variational modulation models (\S~\ref{sec:model}) help achieve instance-level diversity, the noises, sampled regardless of instance styles (\S~\ref{sec:sampling}), can potentially introduce ambiguities during the supervised training (especially for the popular perceptual and feature matching losses), since similar noise samples can possibly correspond to instances of distinct styles. This will affect the effective diversity of generated instances and prohibit the possibility to control the instance styles with certain references.

In light of this, we propose a prior noise remapping step, during which a set of linear perturbation parameters are encoded from given references, to remap the noise samples while preserving the original distribution, in order to provide guidance to embed more meaningful instance diversity in the modulation models. To achieve this, we adopt a noise remapping encoder $\mathcal{E}(\bm{r})$ that translates the reference image $\bm{r}\in\mathbb{R}^{H\times W}$ into four perturbation maps $\{\tilde{\bm{a}}'_\gamma,\tilde{\bm{b}}'_\gamma,\tilde{\bm{a}}'_\beta,\tilde{\bm{b}}'_\beta\}\in\mathbb{R}^{H\times W}$, which are per-pixel linear transformation parameters, including scale $\tilde{\bm{a}}$ and shift $\tilde{\bm{b}}$, for both $\bm{N}_\gamma$ and $\bm{N}_\beta$. A instance aware partial convolution~\cite{harley2017segmentation,liu2018image} is used to avoid information contamination between different instances.
Based on these dense perturbation maps, we apply an instance average pooling layer to each of these maps to get the instance-wise perturbation parameters $\{\tilde{\bm{a}}_\gamma,\tilde{\bm{b}}_\gamma,\tilde{\bm{a}}_\beta,\tilde{\bm{b}}_\beta\}\in\mathbb{R}^{L^p}$. Take $\bm{N}_\gamma$ as an example, for an instance label $l^p\in\mathbb{L}^p$ and its occupying pixels $\bm{x}(l^p)=\{x|\bm{p}[x]=l^p\}$, we have
\begin{equation}
\begin{split}
   \tilde{\bm{a}}_\gamma[l^p]=(\begin{matrix}\sum_{x\in\bm{x}(l^p)}\end{matrix}\tilde{\bm{a}}'_\gamma[x])/|\bm{x}(l^p)|,\\
   \tilde{\bm{b}}_\gamma[l^p]=(\begin{matrix}\sum_{x\in\bm{x}(l^p)}\end{matrix}\tilde{\bm{b}}'_\gamma[x])/|\bm{x}(l^p)|.
\end{split}
\end{equation}
This remapping encoder, together with the main generator, forms a variational autoencoder (VAE)~\cite{kingma2013auto}. The remapped noise samples after perturbation are:
\begin{equation}
   \tilde{\bm{N}}_\gamma[l^p]=\tilde{\bm{a}}_\gamma[l^p]\bm{N}_\gamma[l^p]+\tilde{\bm{b}}_\gamma[l^p],
\end{equation}
where KL-Divergence loss~\cite{kingma2013auto} is used to enforce a same normal distribution $\tilde{\bm{N}}_\gamma\sim\mathcal{N}(0,1)$. These remapped noise samples are used instead of $\bm{N}_\gamma$ during modulation sampling, as described in \S~\ref{sec:sampling}.

During training, the reference $\bm{r}$ is exactly the ground-truth paired image. At test time, while initially sampled noises can be used to achieve random style synthesis by default, as described in \S~\ref{sec:sampling}, it is also allowed to provide $\bm{r}$ as instance references to control the style of the result at the instance level.

\begin{figure*}[tp]
  \centering
  \includegraphics[width=\linewidth]{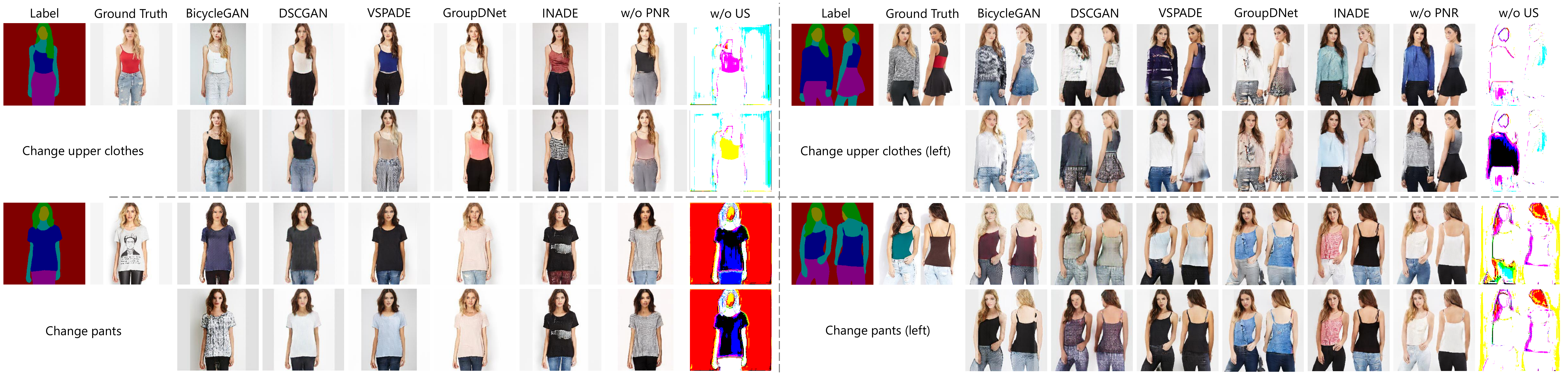}
  \caption{Visual comparison with other multimodal models and two baselines. The results on the left show the performance of class level diversity while the results on the right are for instance level diversity. The first two rows represent the results of different models when changing upper clothes while the last two rows represent the results of changing pants. }
  \label{fig:results_multi}
\end{figure*}
%------------------------------------------------------------------------
\section{Experiments}
\subsection{Implementation Details}
Our INADE generator (Figure~\ref{fig:generator}) follows a similar architecture of the SPADE generator~\cite{park2019semantic}, but with all the SPADE layers replaced by the INADE layers. 
%As shown in Figure~\ref{fig:generator}, it consists of INADE residual blocks and upsampling layers. 
Following SPADE~\cite{park2019semantic}, the overall loss function consists of four loss terms: conditional adversarial loss, feature matching loss~\cite{wang2018high}, perceptual loss~\cite{johnson2016perceptual} and KL-Divergence loss~\cite{kingma2013auto}. Details are provided in the supplementary material.

During training, by default, Adam optimizer~\cite{kingma2015adam} ($\beta_1=0$, $\beta_2=0.9$) is used with fixed epoch number of $200$. The learning rates for the generator and the discriminator are set to $0.0001$ and $0.0004$ respectively, which are gradually decreased to zero after $100$ epochs. Noise $C^0$ has $64$ initialized channels, while the input noise has $256$ channels, same as~\cite{park2019semantic,liu2019learning}. 
All experiments are implemented in PyTorch~\cite{paszke2019pytorch} and conducted on TITAN XP GPUs.

\subsection{Datasets and Metrics}
\noindent\textbf{Datasets.} Experiments are conducted on five popular datasets:  \textit{Cityscapes}~\cite{cordts2016cityscapes}, \textit{ADE20K}~\cite{zhou2017scene}, \textit{CelebAMask-HQ}~\cite{lee2020maskgan,karras2017progressive,liu2015deep}, \textit{DeepFashion}~\cite{liu2016deepfashion}, and \textit{DeepFashion2}. 
\textit{DeepFashion2} is built from \textit{DeepFashion} that each image contains two persons, which is only used for testing. More details can be found in the supplementary material.

\noindent\textbf{Quality Metrics.} To evaluate the result quality, we adopt two types of metrics following~\cite{chen2017photographic,wang2018high,park2019semantic,tan2020semantic}. One is the Fr\'echet Inception Distance (FID)~\cite{heusel2017gans}, which measures the distance of distributions between results and real images.

The other one is semantic-segmentation-based, which evaluates the semantic segmentation accuracy on the results by comparing the predicted masks with the groundtruth layouts on both mean Intersection-over-Union (mIoU) and pixel accuracy (accu). State-of-the-art pretrained segmentation models are used for different datasets: DRN~\cite{yu2017dilated,github-drn} for \textit{Cityscapes}, UperNet101~\cite{xiao2018unified,github-upernet} for \textit{ADE20k}, and UNet~\cite{ronneberger2015u,github-celebamask} for \textit{CelebAMask-HQ}. For \textit{DeepFashion}, we use the same UNet-based network but train the model by ourselves. For fair evaluation, we run the model for $10$ times and report the average scores when noise input is required.

\noindent\textbf{Diversity Metrics.} To evaluate the diversity of the results, we adopt the LPIPS metric as proposed by~\cite{zhang2018unreasonable,github-lpips}. Similar to~\cite{zhu2020semantically}, we also adopt two metrics to measure the semantic-level diversity (mCSD and mOCD) and expand them to instance level (mISD and mOID). More details can be found in the supplementary material.

\noindent\textbf{Subjective Metrics.} We conduct human evaluations to assess both the quality and the diversity of the methods. For quality, we ask the volunteers to select the most realistic one among the results generated by different methods on the same input. mHE (mean Human Evaluation) denotes the percentage of results being selected for each method.

For diversity, we expand the metric proposed by~\cite{zhu2020semantically} to the instance level. A pair of results, with one random semantic class or instance manipulated, are given to volunteers. The percentage of pairs that are judged to be different in only one area represents the human evaluation, namely SHE (Semantic Human Evaluation) and IHE (Instance Human Evaluation). We invite $20$ volunteers for evaluation and the evaluated number of images is $20$.

\begin{table*}[t]
\setlength{\tabcolsep}{1.6mm}
    \centering
    \caption{Comparison with other multimodal methods on diversity. mHE, SHE and IHE are aforementioned metrics.
    $\uparrow$ and $\downarrow$ represent the higher the better and the lower the better. \textbf{Bold} and \underline{underlined} numbers are the best and the second best of each metric, respectively.}
    \small
    \begin{tabular}{c|c|c|c|c|c|c|c|c|c|c|c|c}
    \hline
        \multirow{2}{*}{Methods} & \multicolumn{6}{c|}{DeepFashion} & \multicolumn{6}{c}{DeepFashion2}\\
        \cline{2-13}
         & FID $\downarrow$ & LPIPS $\uparrow$ & mCSD $\uparrow$ & mOCD $\downarrow$ & mHE $\uparrow$ & SHE $\uparrow$ & FID $\downarrow$ & LPIPS $\uparrow$ & mISD $\uparrow$ & mOID $\downarrow$ & mHE $\uparrow$ & IHE $\uparrow$\\
        \hline
        BicycleGAN & 31.10 & \textbf{0.225} & \underline{0.0465} & 0.2014 & 0.0 & 4.5 & 33.46 & \underline{0.286} & \underline{0.0500} & 0.2456 & 0.0 & 2,5\\
        \hline
        DSCGAN & 29.79 & 0.146 & 0.0404 & 0.1218 & 0.0 & 9.3 & 48.64 & 0.199 & 0.0433 & 0.1633 & 0.0 & 4.8\\
        \hline
        VSPADE & 11.11 & 0.197 & 0.0450 & 0.1665 & 7.5 & 6.5 & \underline{22.29} & 0.222 & 0.0390 & 0.1780 & \underline{43.7} & 3.3\\
        \hline
        GroupDNet & \textbf{9.72} & 0.222 & 0.0453 & \textbf{0.0077} & \underline{40.0} & \underline{86.0} & 22.81 & 0.281 & 0.0434 & \underline{0.0303} & 8.8 & \underline{9.3}\\
        \hline
        INADE & \underline{9.97} & \textbf{0.225} & \textbf{0.0511} & \underline{0.0161} & \textbf{52.5} & \textbf{88.3} & \textbf{18.18} & \textbf{0.319} & \textbf{0.0580} & \textbf{0.0187} & \textbf{47.5} & \textbf{82.8}\\
        \hline
        \hline
        w/o PNR & 12.09 & 0.184 & 0.0289 & 0.0138 & - & - & 20.76 & 0.243 & 0.0291 & 0.0189 & - & -\\
        \hline
        w/o US & 248.33 & 0.624 & 0.0730 & 0.0370 & - & - & 265.63 & 0.633 & 0.0748 & 0.0296 & - & -\\
        \hline
    \end{tabular}
    \label{tab:comp_diversity}
\end{table*}{}

\begin{table*}[t]
\setlength{\tabcolsep}{1.6mm}
    \centering
    \caption{Comparison with SOTA methods on result quality. All the numbers are collected by running the evaluation on our machine. Here $\mathbf{M}$, $\mathbf{A}$, $\mathbf{F}$, and $\mathbf{L}$ represent mIoU, accu, FID, and LPIPS, respectively. Note that the $\mathbf{L}$ score of SEAN is almost zero even with noise input.}
    %\small
    \footnotesize
    \begin{tabular}{c|c|c|c|c|c|c|c|c|c|c|c|c|c|c|c|c}
    \hline
        \multirow{2}{*}{Methods} & \multicolumn{4}{c|}{Cityscapes} & \multicolumn{4}{c|}{ADE20K} & \multicolumn{4}{c|}{CelebAMask-HQ} & \multicolumn{4}{c}{DeepFashion}\\
        \cline{2-17}
         & $\mathbf{M}$ $\uparrow$ & $\mathbf{A}$ $\uparrow$ & $\mathbf{F}$ $\downarrow$ & $\mathbf{L}$ $\uparrow$ & $\mathbf{M}$ $\uparrow$ & $\mathbf{A}$ $\uparrow$ & $\mathbf{F}$ $\downarrow$ & $\mathbf{L}$ $\uparrow$ & $\mathbf{M}$ $\uparrow$ & $\mathbf{A}$ $\uparrow$ & $\mathbf{F}$ $\downarrow$ & $\mathbf{L}$ $\uparrow$ & $\mathbf{M}$ $\uparrow$ & $\mathbf{A}$ $\uparrow$ & $\mathbf{F}$ $\downarrow$ & $\mathbf{L}$ $\uparrow$\\
        \hline
        SPADE & \textbf{61.38} & \underline{93.26} & 51.98 & 0 & \textbf{36.28} & \underline{78.13} & 29.79 & 0 & 75.22 & 94.76 & 31.40 & 0 & \textbf{76.76} & \textbf{97.65} & 11.22 & 0\\
        \hline
        pix2pixHD & 60.50 & 93.06 & 66.04 & 0 & 27.27 & 72.61 & 45.87 & 0 & \underline{76.11} & \textbf{95.67} & 36.95 & 0 & 73.99 & 97.02 & 15.27 & 0\\
        \hline
        CLADE & 60.44 & \textbf{93.42} & 50.62 & 0 & \underline{35.43} & 77.37 & 30.48 & 0 & 75.37 & 95.05 & 33.54 & 0 & 75.63 & 97.33 & 12.76 & 0\\
        \hline
        SEAN & 56.22 & 92.28 & 50.43 & 0 & 32.65 & 76.58 & \textbf{28.11} & 0 & 75.94 & 95.03 & \underline{24.30} & 0 & \underline{76.28} & 97.46 & \textbf{7.37} & 0\\
        \hline
        BicycleGAN & 30.47 & 78.26 & 59.87 & 0.122 & 5.33 & 42.68 & 77.49 & \textbf{0.443} & 65.98 & 89.77 & 35.73 & \underline{0.362} & 73.09 & 96.75 & 31.10 & \textbf{0.225}\\
        \hline
        DSCGAN & 43.70 & 87.80 & 50.84 & \underline{0.216} & 8.07 & 58.10 & 82.30 & 0.324 & 75.98 & 95.08 & 52.83 & 0.198 & 75.92 & 96.97 & 29.79 & 0.146\\
        \hline
        GroupDNet & 59.20 & 92.78 & \underline{41.12} & 0.073 & 26.09 & 73.07 & 39.11 & 0.177 & \textbf{76.13} & \underline{95.21} & 29.39 & 0.309 & 76.19 & \underline{97.48} & \underline{9.72} & 0.222\\
        \hline
        INADE & \underline{61.02} & 93.16 & \textbf{38.04} & \textbf{0.248} & 34.96 & \textbf{78.51} & \underline{29.60} & \underline{0.400} & 74.08 & 94.31 & \textbf{22.55} & \textbf{0.365} & 76.27 & 97.44 & 9.96 & \textbf{0.225}\\
        \hline
    \end{tabular}
    \label{tab:comp_quality}
\end{table*}{}

\begin{table*}[t]
\setlength{\tabcolsep}{1.8mm}
    \centering
    \caption{Comparison with other semantic image synthesis methods on model complexity and efficiency. All the numbers are collected by running the evaluation on Titan XP. Here $\mathbf{P}$ and $\mathbf{T}$ denote the number of generator parameters and inference run time, respectively.}
    %\small
    \footnotesize
    \begin{tabular}{c|c|c|c|c|c|c|c|c|c|c|c|c}
    \hline
        \multirow{2}{*}{Methods} & \multicolumn{3}{c|}{Cityscapes} & \multicolumn{3}{c|}{ADE20K} & \multicolumn{3}{c|}{CelebAMask-HQ} & \multicolumn{3}{c}{DeepFashion}\\
        \cline{2-13}
         & $\mathbf{P}$ (M) & FLOPs (G) & $\mathbf{T}$ (s) & $\mathbf{P}$ (M) & FLOPs (G) & $\mathbf{T}$ (s) & $\mathbf{P}$ (M) & FLOPs (G) & $\mathbf{T}$ (s) & $\mathbf{P}$ (M) & FLOPs (G) & $\mathbf{T}$ (s)\\
        \hline
        SPADE & 93.05 & 281.54 & 0.065 & 96.50 & 181.30 & 0.042 & 92.54 & 141.32 & 0.035 & 92.21 & 137.99 & 0.032\\
        \hline
        pix2pixHD & 182.53 & 151.32 & 0.038 & 182.90 & 99.30 & 0.041 & 182.47 & 72.17 & 0.023 & 182.44 & 69.91 & 0.020\\
        \hline
        CLADE & 67.90 & 75.54 & 0.035 & 71.40 & 42.20 & 0.024 & 67.32 & 42.15 & 0.022 & 66.98 & 42.15 & 0.019\\
        \hline
        SEAN & 330.41 & 681.75 & 0.507 & - & - & - & 266.90 & 346.27 & 0.165 & 223.23 & 342.89 & 0.135\\
        \hline
        BicycleGAN & 54.80 & 18.40 & 0.011 & 54.80 & 18.40 & 0.006 & 54.80 & 18.40 & 0.006 & 54.80 & 18.40 & 0.006\\
        \hline
        DSCGAN & 54.00 & 18.14 & 0.018 & 54.00 & 18.14 & 0.010 & 54.00 & 18.14 & 0.010 & 54.00 & 18.14 & 0.010\\
        \hline
        GroupDNet & 76.50 & 463.61 & 0.224 & 68.33 & 383.00 & 0.088 & 145.29 & 225.53 & 0.090 & 96.32 & 291.61 & 0.062\\
        \hline
        INADE & 77.39 & 75.25 & 0.048 & 90.89 & 42.19 & 0.035 & 85.12 & 42.18 & 0.030 & 84.63 & 42.92 & 0.026\\
        \hline
    \end{tabular}
    \label{tab:comp_complexity}
\end{table*}{}

\begin{figure}[tp]
  \centering
  \includegraphics[width=\linewidth]{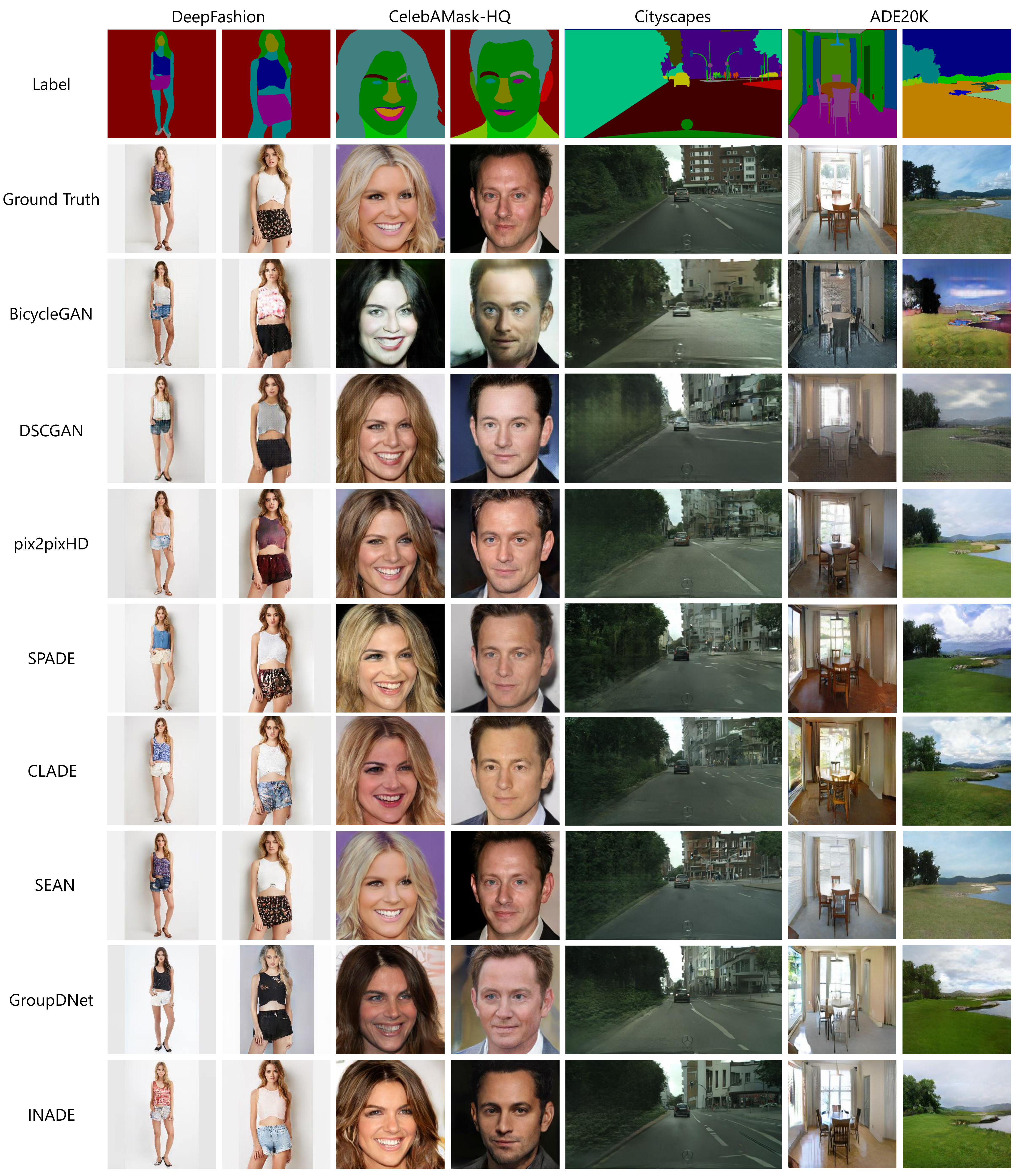}
  \caption{Qualitative comparison with the state-of-the-art semantic image synthesis methods on four datasets: DeepFashion, CelebAMask-HQ, Cityscapes and ADE20K.}
  \label{fig:results_stoa}
\end{figure}

\begin{figure}[tp]
  \centering
  \includegraphics[width=\linewidth]{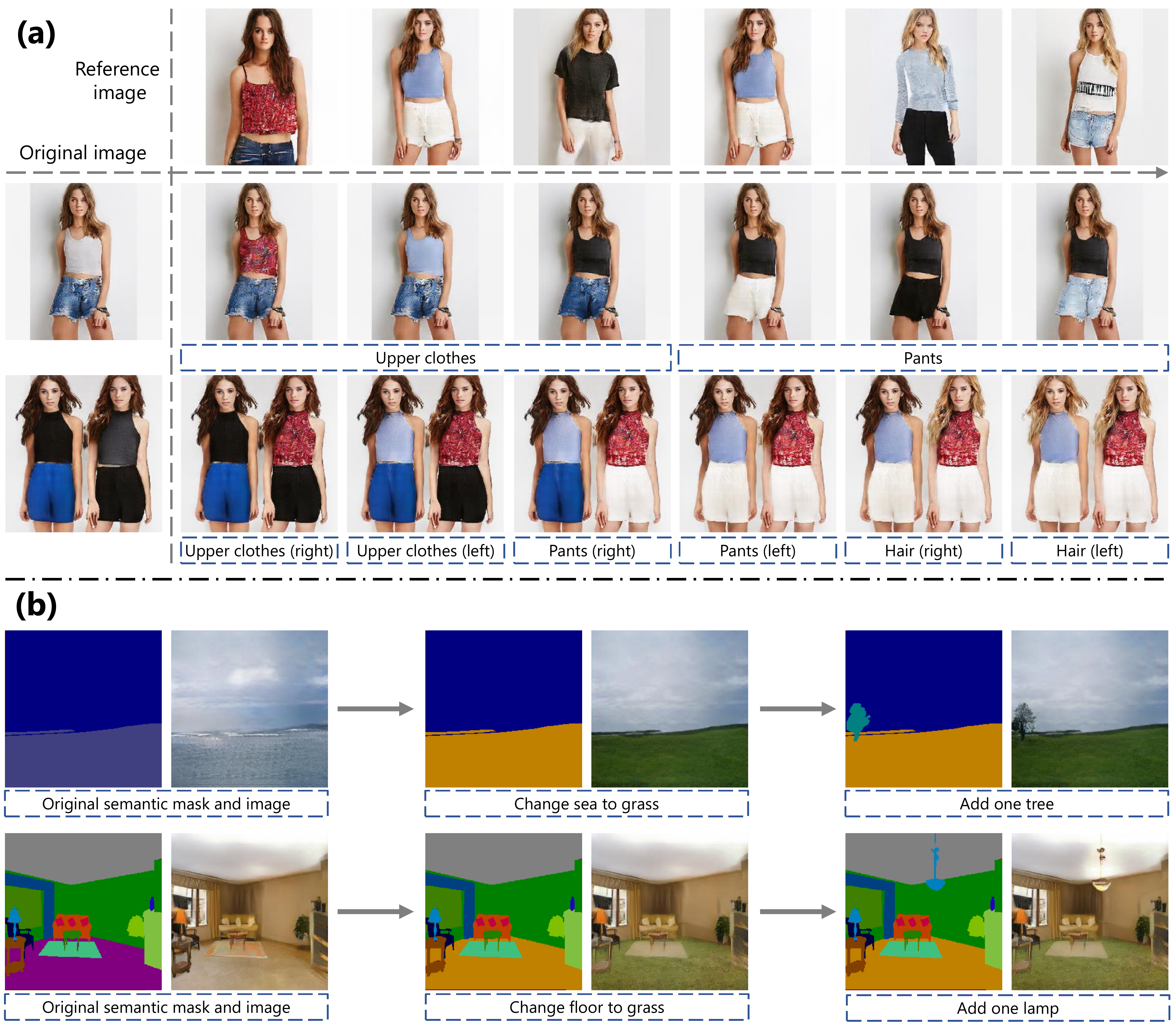}
  \caption{Exemplar applications of the proposed INADE. (a) Results of our method for reference appearance editing. From left to right, we change the appearance of part of the target image based on the reference image (the changed instance is indicated by the text in the blue dotted box). (b) Application of our method for semantic manipulation. The text in the blue dotted box indicates what is edited each time.}
  \label{fig:apps}
\end{figure}

\subsection{Quantitative and Qualitative Comparisons}
We compare INADE with several SOTA works, including quality-oriented (pix2pixHD~\cite{wang2018high}, SPADE~\cite{park2019semantic}, CLADE~\cite{tan2020rethinking,tan2020semantic} and SEAN~\cite{zhu2020sean})
and diversity-oriented (BicycleGAN~\cite{zhu2017toward}, DSCGAN~\cite{yang2018diversity}, and GroupDNet~\cite{zhu2020semantically}) methods. For a fair comparison, we directly use the pretrained models provided by the authors when available, otherwise train the models by ourselves using the codes and settings provided by the authors. For SPADE, which has the strategy for multimodal synthesis, we train the model with that extra encoder but ignore it when testing its diversity performance (namely VSPADE). Reference images are not allowed for any of the methods during testing except for SEAN which requires the reference input.

\subsubsection{Multimodal Image Synthesis}
Several methods that support multimodal image synthesis are compared to demonstrate the superior diversity performance of INADE. In additional, we also compare with two ablation baselines: w/o \textbf{US} (without \textbf{U}nified \textbf{S}ampling, \S\ref{sec:sampling}) and w/o \textbf{PNR} (without \textbf{P}rior \textbf{N}oise \textbf{R}emapping, \S\ref{sec:noise}). w/o \textbf{US} means that the noise for each INADE layers are sampled independently, while w/o \textbf{PNR} means that the PNR is not used during training.
The quantitative results on the \textit{DeepFashion} dataset are summarized in Table~\ref{tab:comp_diversity}.

In general, INADE achieves superior performance regarding both quality and diversity compared to previous methods. 
For single-subject images in the \textit{DeepFashion} dataset, our method exhibits better performance than BicycleGAN, DSCGAN, and VSPADE in terms of FID, and is comparable to GroupDNet. While for multiple-subject images (\textit{DeepFashion2}), INADE shows the lowest FID.

In terms of diversity, our method achieves the best scores on metrics including overall measurement (LPIPS) and semantic-level/instance-level metrics (mCSD/mISD). As for mOCD, methods only support image-level diversity, such as BicycleGAN, DSCAN, and VSPADE, produce much more unwanted changes outside the instance area. Although both being relatively low, INADE is higher than GroupDNet, which is because GroupDNet uses group convolution to prevent premature fusion between features of different classes, while INADE uses conventional convolution for more consistent combinations. 
For mOID, as the only method that supports instance-wise control, INADE easily gets the best score. More analysis of mOID and mOCD can be found in the supplementary material.

As for subjective evaluations, our method also outperforms others on both semantic- and instance-level cases. 

Compared to the two ablation baselines, we find that both prior noise remapping and unified sampling play indispensable roles in our method. Removing prior noise remapping (w/o \textbf{PNR}) leads to ambiguities during the supervised training, which seriously affects the quality and diversity of synthesized results. Independently sampling (w/o \textbf{US}) for each normalization destroys the consistency of information and significantly degenerates the generation result.

The qualitative comparisons are shown in Figure~\ref{fig:results_multi}. Although all methods support multimodal synthesis, the quality by BicycleGAN and DSCGAN is not satisfactory. VSPADE achieves good visual quality, but does not support semantic- or instance-level control. GroupDNet is capable of changing the appearance of a specific semantic class (results on the left), but tends to generate identical style for different cloth instances (results on the right). On the contrary, INADE supports both fine-grained multimodal controls with high visual fidelity. As for the ablation baselines, we notice that removing \textbf{PNR} significantly decreases the quality, while the whole task fails without \textbf{US}. 

\subsubsection{Semantic Image Synthesis}
The quantitative comparisons against semantic image synthesis methods are summarized in Table~\ref{tab:comp_quality}.

Compared to methods that don't support multimodal (i.e. 0 LPIPS), especially SEAN which has additional reference image input, INADE has an advantage or near the best on almost all metrics. It seems that SPADE has slight advantage in segmentation metrics, but INADE still shows its overall superiority when considering FID score and visual results.

Compared to existing multimodal methods, INADE leads all metrics. In terms of quality (e.g. mIoU, acc, and FID), BicycleGAN and DSCGAN are much lower than ours on all datasets. GroupDNet achieves similar or slightly better performance on \textit{CelebAMask-HQ} and \textit{DeepFashion} datasets, but has a significant gap on more complicated scenes such as \textit{Cityscapes} and \textit{ADE20K}. This demonstrates the superiority of the proposed method on synthesizing complex scenes. The LPIPS score shows that all these methods are able to generate multimodal images to some extent. 
BicycleGAN gets higher scores than ours on some datasets, but is not able to do high-quality synthesis.
GroupDNet shows good performance on person-related tasks, but falls into strong bias when dealing with complex scenes which greatly restricts its performance. Therefore, considering both quality and diversity, our method achieves the best overall performance.

Qualitative comparisons on these four datasets are shown in Figure~\ref{fig:results_stoa}. In general, the images generated by INADE are more realistic than others on various datasets, which is consistent with the quantitative results.

\subsubsection{Computational and Model Complexity}
In this section, we analyze the computational and model overhead of different methods. The quantitative results (generator networks only) are summarized in Table~\ref{tab:comp_complexity}.

BicycleGAN and DSCGAN share a similar small network architecture with the least parameters, FLOPs (floating-point operations per second), and run-time cost. However, the quality of the synthesized images is far from satisfactory. CLADE seems to get a good trade-off between performance and efficiency, but still falls short of INADE in terms of overall performance and functionality.
Compared to all other methods, INADE achieves the smallest network (parameters and FLOPs), as well as one of the fastest run-time performance.
Specifically, compared with GroupDNet, the only method that can achieve semantic-level diversity, our method provides control over both semantic and instance levels with much less overhead, introducing $82\%\sim89\%$ fewer FLOPs and $59\%\sim79\%$ less inference time compared with GroupDNet.

\subsection{Applications}
Thanks to its superior capability of controllable diverse synthesis, INADE can be used in many image editing applications. Here we show two examples as follows.

\noindent\textbf{Reference appearance editing.} With the noise remapping mechanism described in \S~\ref{sec:noise}, we can extract the instance-wise style from an arbitrary reference image. This makes it possible for INADE to perform reference-based editing to different parts of an image at the instance level. As shown in Figure~\ref{fig:apps} (a), we can change the appearance of hairstyles, tops, and pants to match the reference. 

\noindent\textbf{Semantic manipulation.} Similar to most existing semantic image synthesis methods, INADE also supports semantic manipulation. We show some examples in Figure~\ref{fig:apps} (b), such as changing the semantic class to an object, or insert a new semantic object into the image. And more creative editing results can be achieved by modifying the semantic mask and the instance map.

\section{Conclusion}
In this paper, we focus on multimodal image synthesis and propose a novel diverse semantic image synthesis method based on instance-aware conditional normalization. Different from previous works, we learn the class-wise probability distributions and perform instance-wise stochastic sampling to generate the per-instance modulation parameters. Our method improves the network's ability to model semantic categories and make it easier to synthesize diverse images at semantic- or instance-level without scarifying the visual fidelity.

{\small
\bibliographystyle{ieee_fullname}
\bibliography{egbib}
}
\newpage

\section{Additional Implementation details}
\begin{figure}[tp]
  \centering
  \includegraphics[width=\linewidth]{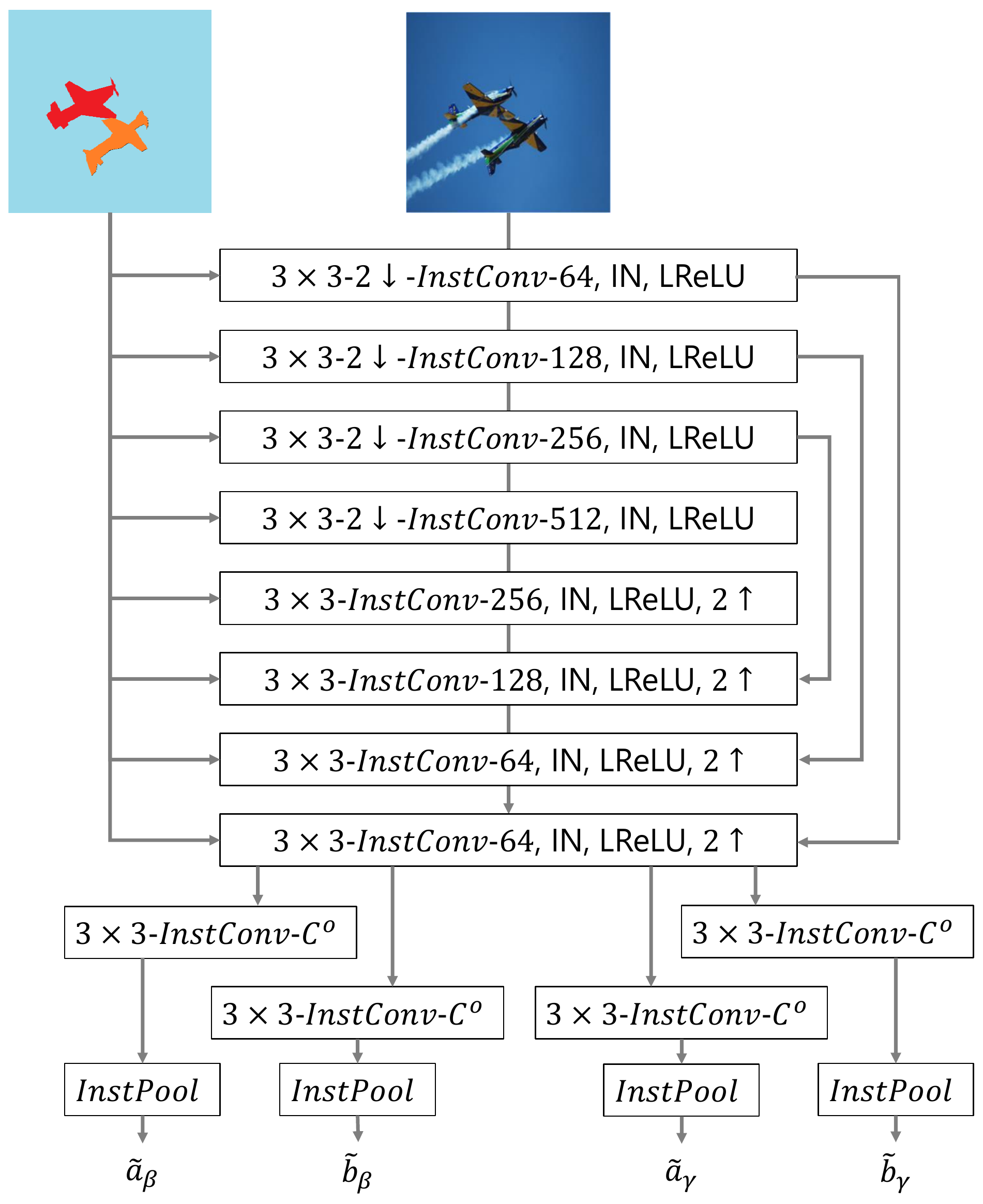}
  \caption{Architecture of our encoder network. We use UNet~\cite{ronneberger2015u} based network to extract the features with the same resolution of input image, and then obtain the $\{\tilde{\bm{a}}_\gamma,\tilde{\bm{b}}_\gamma,\tilde{\bm{a}}_\beta,\tilde{\bm{b}}_\beta\}$ through independent instance partial convolution (InstConv) and instance average pooling (InstPool).}
  \label{fig:encoder}
\end{figure}

\begin{figure}[tp]
  \centering
  \includegraphics[width=\linewidth]{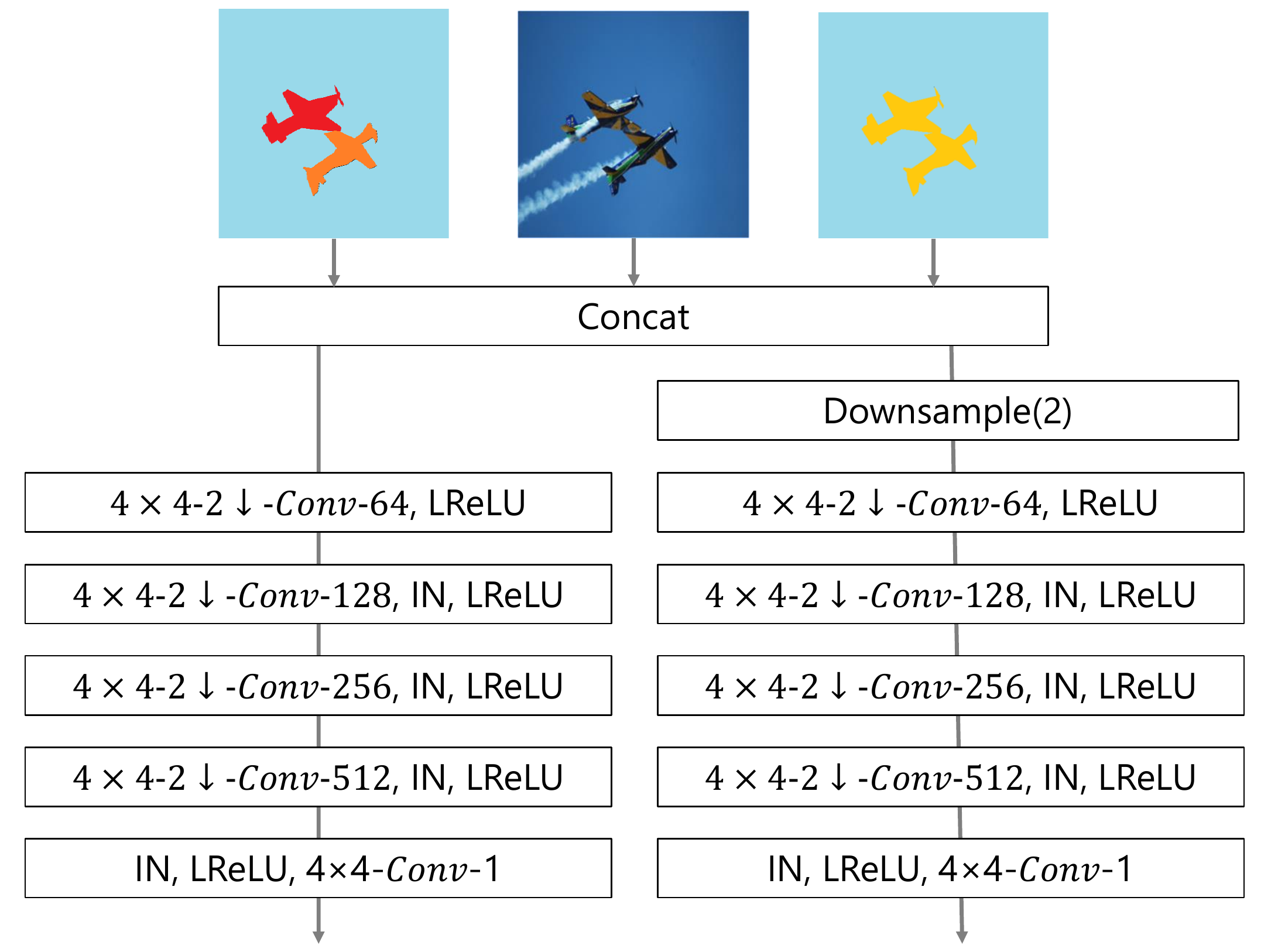}
  \caption{The discriminator of our method is based on the PatchGAN~\cite{isola2017image}. It takes the concatenation the segmentation map, instance map and the image as input.}
  \label{fig:discriminator}
\end{figure}

\noindent\textbf{Network architectures.} Here we give detailed network designs for each part. Figure~\ref{fig:encoder} shows the architecture of our encoder. We use instance partial convolution and instance average pooling to get the parameters of each instance independently. The architecture of generator network is shown in Figure~\ref{fig:generator}. The synthesis process starts with a random noise and goes through a series of the proposed INADE ResBIKs. Since the training is carried out on multiple GPUs, the batch normalization layer in INADE adopts the synchronous version. We use a multi-scale PathGAN~\cite{isola2017image} based discriminator whose architecture is shown in Figure~\ref{fig:discriminator}.

\noindent\textbf{Loss function.} The loss function we adopted consists of four components:

\textit{Conditional adversarial loss.} Let $\mathcal{E}$ be the prior noise remapping, G be the INADE generator, D be the discriminator, $\bm{m}$ be a given semantic mask, $\bm{o}$ and $\bm{p}$ be the corresponding image and instance map. The conditional adversarial loss built with hinge loss is formulated as:
\begin{equation}
\begin{split}
\mathcal{L}_{GAN}(\mathcal{E}, G, D) = \mathbb{E}[max(0,1-D(\bm{o},\bm{m},\bm{p}))] \\+ \mathbb{E}[max(0,1+D(G(\mathcal{E}(\bm{o},\bm{p}),\bm{m},\bm{p}),\bm{m},\bm{p}))].
\end{split}
\end{equation}

\textit{Feature matching loss.} Let $D_i$ and $N_i$ be the output feature maps and the number of elements of the $i$-the layer of D respectively, $S_D$ and $E_D$ be the start number of layer for loss calculation and total number layers in D respectively. The feature matching loss is denoted as:
\begin{equation}
\begin{split}
    \mathcal{L}_{F}=\mathbb{E}\sum_{i=S_D}^{E_D} \frac{1}{N_i}[\Vert D_i(\bm{o},\bm{m},\bm{p})-\\
    D_i(G(\mathcal{E}(\bm{o},\bm{p}),\bm{m},\bm{p}),\bm{m},\bm{p}))\Vert_1].
\end{split}
\end{equation}
To reduce the ambiguity, we only use high-level features and set $S_D$ to $3$.

\textit{Perceptual loss.} Let $V_i$ and $M_i$ be the output feature maps and the number of elements of the $i$-the layer of VGG network respectively, $S_V$ and $E_V$ be the start number of layer for loss calculation and total number layers in VGG network respectively. The perceptual loss is denoted as:
\begin{equation}
\begin{split}
    \mathcal{L}_{P}=\mathbb{E}\sum_{i=S_V}^{E_V} \frac{1}{M_i}[\Vert V_i(\bm{o})-
    V_i(G(\mathcal{E}(\bm{o},\bm{p}),\bm{m},\bm{p}))\Vert_1].
\end{split}
\end{equation}
Similar to feature matching loss, we only use high-level features and set $S_D$ to $3$.

\textit{KL-Divergence loss.} Let $q_\beta(z|\bm{o},\bm{p})$ and $q_\gamma(z|\bm{o},\bm{p})$ be the variational distribution of $\bm{N}_\gamma$ and $\bm{N}_\beta$ respectively. $p(z)$ be a standard Gaussian distribution. The KL-Divergence loss is denoted as:
\begin{equation}
    \mathcal{L}_{KL}=0.5\times(\mathcal{D}(q_\beta(z|\bm{o},\bm{p}) \Vert p(z)) + \mathcal{D}(q_\gamma(z|\bm{o},\bm{p}) \Vert p(z))).
\end{equation}

The overall loss is made up of the above-mentioned loss terms as:
\begin{equation}
\begin{split}
    \min_{\mathcal{E}, G}(\max_{D}(\mathcal{L}_{GAN})+\lambda_{1} \mathcal{L}_{F}+ \lambda_{2} \mathcal{L}_{P}+\lambda_{3} \mathcal{L}_{KL}),
\end{split}
\end{equation}
Following SPADE, We set $\lambda_{1}=10, \lambda_{2}=10, \lambda_{3}=0.05$. 

\section{Details of Datasets}
The details about each dataset are described as follows:
\begin{itemize}
    \item \textit{Cityscapes} dataset~\cite{cordts2016cityscapes} is a widely used dataset for semantic image synthesis~\cite{wang2018video,qi2018semi,wang2018high}. The high-resolution images with fine semantic and instance annotations are taken from street scenes of German cities. There are 2,975 training images and 500 validation images. The number of annotated semantic classes is 35. 
    \item \textit{ADE20K} dataset~\cite{zhou2017scene} consists of 25,210 images (20,210 for training, 2,000 for validation and 3,000 for testing). The images in \textit{ADE20K} dataset cover a wide range of scenes and object categories, including a total of 150 object and stuff classes. 
    \item \textit{CelebAMask-HQ} dataset~\cite{lee2020maskgan,karras2017progressive,liu2015deep} is based on CelebAHQ face imgae dataset. It contains of 28,000 training images and 2,000 validation images with 19 different semantic classes. 
    \item \textit{DeepFashion} dataset~\cite{liu2016deepfashion} contains 52,712 person images with fashion clothes. We use the processed dataset provided by GroupDNet~\cite{zhu2020semantically} which consists of 30,000 training images and 2,247 validation images. There are 8 different semantic classes.
    \item \textit{DeepFashion2} dataset is built from DeepFashion. We combine two adjacent images to generate the images containing two persons. The new semantic mask and the instance map are also derived from the corresponding two semantic masks. This dataset is only used to evaluate the performance of models trained on DeepFashion dataset in terms of instance level diversity.
\end{itemize}
In these datasets, \textit{Cityscapes} and \textit{DeepFashion2} have semantic and instance annotations, while the rest have only semantic annotations. In our experiment, the resolution of images is $256\times 256$ except that Cityscapes dataset is $256\times 512$.

\section{Details of Diversity Metrics}
We adopt the LPIPS~\cite{zhang2018unreasonable,github-lpips} to evaluate the overall diversity of the results. Specifically, we generate $10$ groups of images or evaluation with randomly sampled noise, and calculate the diversity score between $2$ random groups at a time. A total of $10$ scores are calculated, and we measure the mean of these scores to reduce the potential fluctuation caused by random sampling.

To evaluate the instance-level diversity, we expand the metrics proposed by~\cite{zhu2020semantically}, called mean \textit{Instance-Specific Diversity} (mISD) and mean \textit{Other-Instances Diversity} (mOID), which represent the degree of change inside and outside the instance region when being manipulated. Specially, we generate several images by changing sampled noise for specified instance while keeping the noise for others unchanged. Then, the similarity inside and outside the instance region between these images makes up the mISD and mOID metrics. For datasets which have no instance annotations, these metrics degenerate to semantic level (mean \textit{Class-Specific Diversity} (mCSD) and mean \textit{Other-Classes Diversity} (mOCD)) which are the same with~\cite{zhu2020semantically}. A high diversity inside the instance area (high mISD), as well as a low outside diversity (low mOID), are desired.

\section{Additional ablation study}
Here we give the additional ablation study for $C^0$ which represents the length of the initial sampling. Intuitively, the longer the sampling length is, the higher the diversity of the synthesized image will be. We conduct experiments on the \textit{Cityscapes} and \textit{CelebAMask-HQ} datasets, which include complex street scenes and delicate facial images. As summarized in Table~\ref{tab:ablation}, we compare the default setting ($C^o=64$) with two variant settings: a shorter sampling length ($C^o=8$, INADE-8) and a longer sampling length ($C^o=128$, INADE-128). We find that INADE-8 shows the lower LPIPS score than INADE, while IANDE-128 correspondingly gets the highest score in this metric. And the model with the default setting (INADE) gets the best scores in terms of quality metrics. In our understanding, a short sampling length (e.g. 8) may limit the information capacity, thus reducing the generation quality (low scores of mIoU, acc and FID) and diversity (low score of LPIPS). In contrast, a longer sampling length (e.g. 128) can increase the diversity of the synthesized image (high score of LPIPS), but also increases the difficulty of high-quality image generation (low scores of mIoU, acc and FID). 

In terms of model parameters, FLOPs and run time, INADE-8 is best, but the advantage is not obvious compared with INADE and INADE-128. Based on the above results, we set $C^o=64$ on different datasets by default. 

\section{Additional results}
\begin{figure*}[tp]
  \centering
  \includegraphics[width=\linewidth]{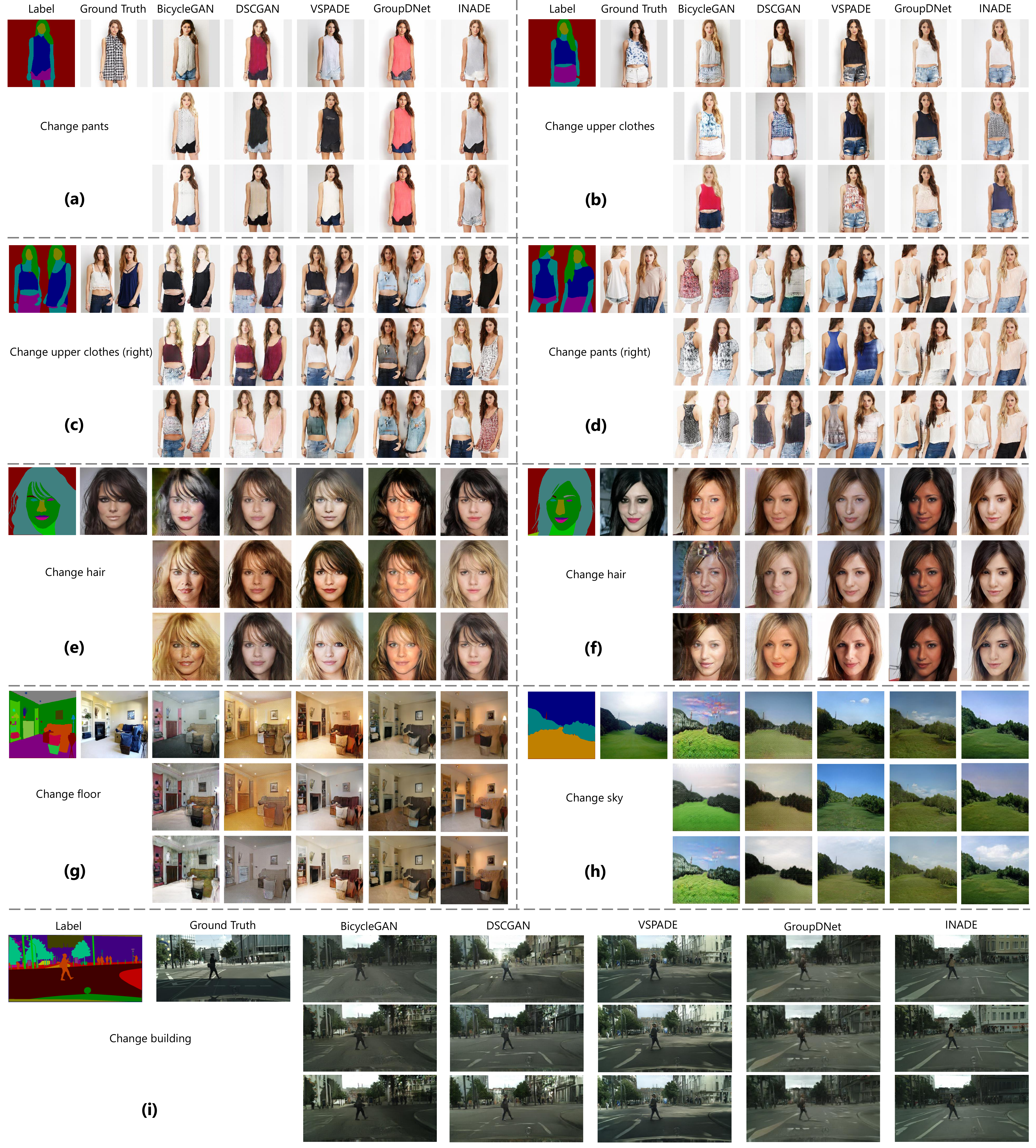}
  \caption{Multi-modal comparison of our INADE with previous state-of-the-art methods on DeepFashion (a-b), DeepFashion2 (c-d), CelebAMask-HQ (e-f), ADE20K (g-h) and Cityscapes (i) datasets.}
  \label{fig:sup_random}
\end{figure*}

In Figure~\ref{fig:sup_random}, we show more multi-modal qualitative results on different datasets that only change one specified class or instance. The conclusions are basically the same as we mention in the main submission. BicycleGAN, DSCGAN and VSPADE show the global style controllabitity, GroupDNet expands it to semantic level, while the synthesis results of our method can be controlled at both the semantic level and instance level. We notice that in some results, when we change one part, other parts slightly change as well, which is also mentioned in GroupDNet~\cite{zhu2020semantically}. In fact, this is reasonable in some cases to 
increase the generation fidelity. For example, as shown in Figure~\ref{fig:sup_random} (h), the lighting often changes with the sky, if the appearance of the grass is totally unchanged, the final generated image will look unnatural to some extent. Therefore, though the metric mOCD (or mOID) may be a good indication of semantic/instance-level controllability, a slightly high mOCD or mOID do not represent worse quality. In other words, we do not expect them to be zero in real applications. 

\begin{figure*}[tp]
  \centering
  \includegraphics[width=\linewidth]{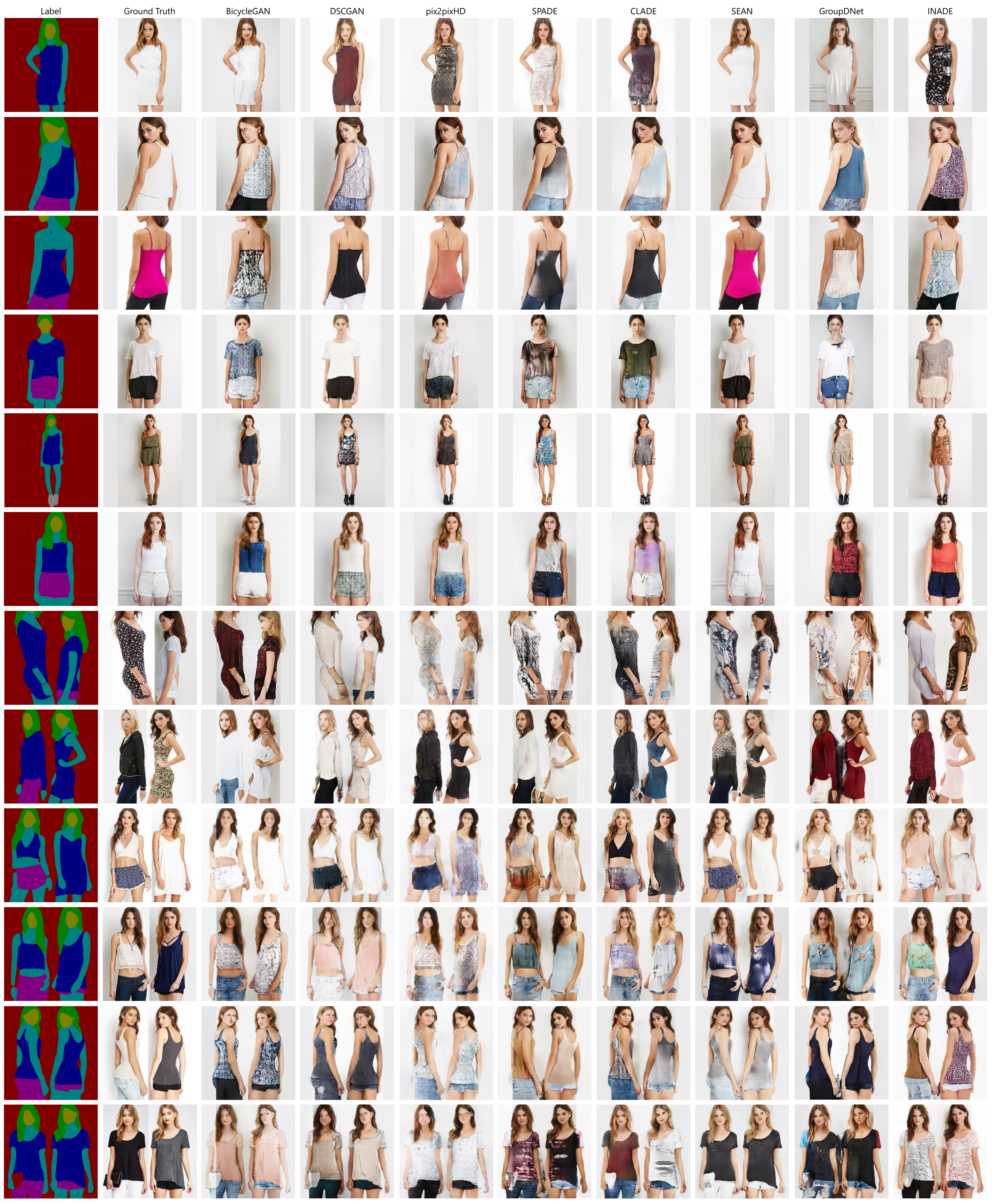}
  \caption{Qualitative comparison of our INADE with previous state-of-the-art methods on the DeepFashion and DeepFashion2 datasets.}
  \label{fig:sup_fashion}
\end{figure*}

\begin{figure*}[tp]
  \centering
  \includegraphics[width=\linewidth]{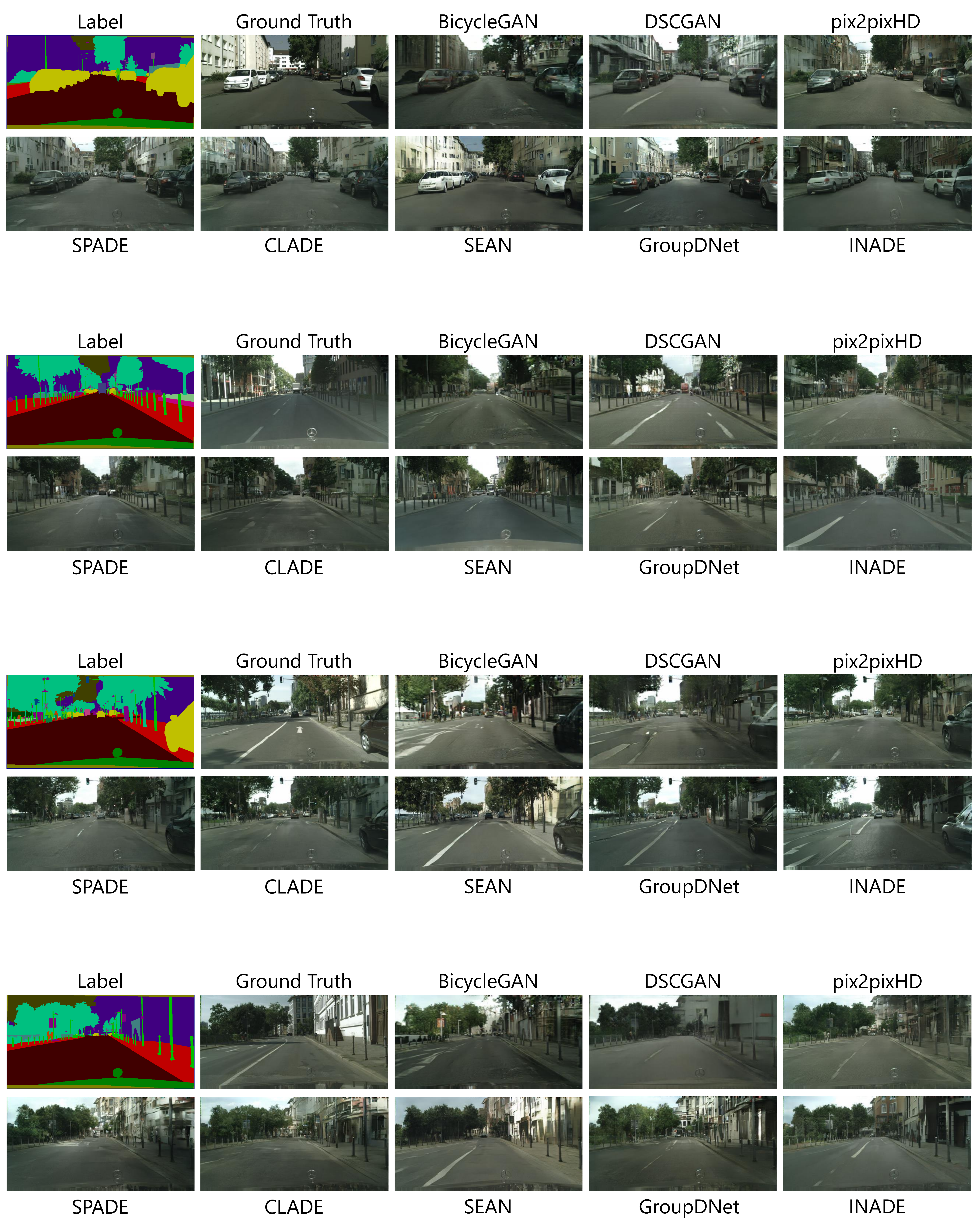}
  \caption{Qualitative comparison of our INADE with previous state-of-the-art methods on the Cityscapes dataset.}
  \label{fig:sup_city}
\end{figure*}

\begin{figure*}[tp]
  \centering
  \includegraphics[width=\linewidth]{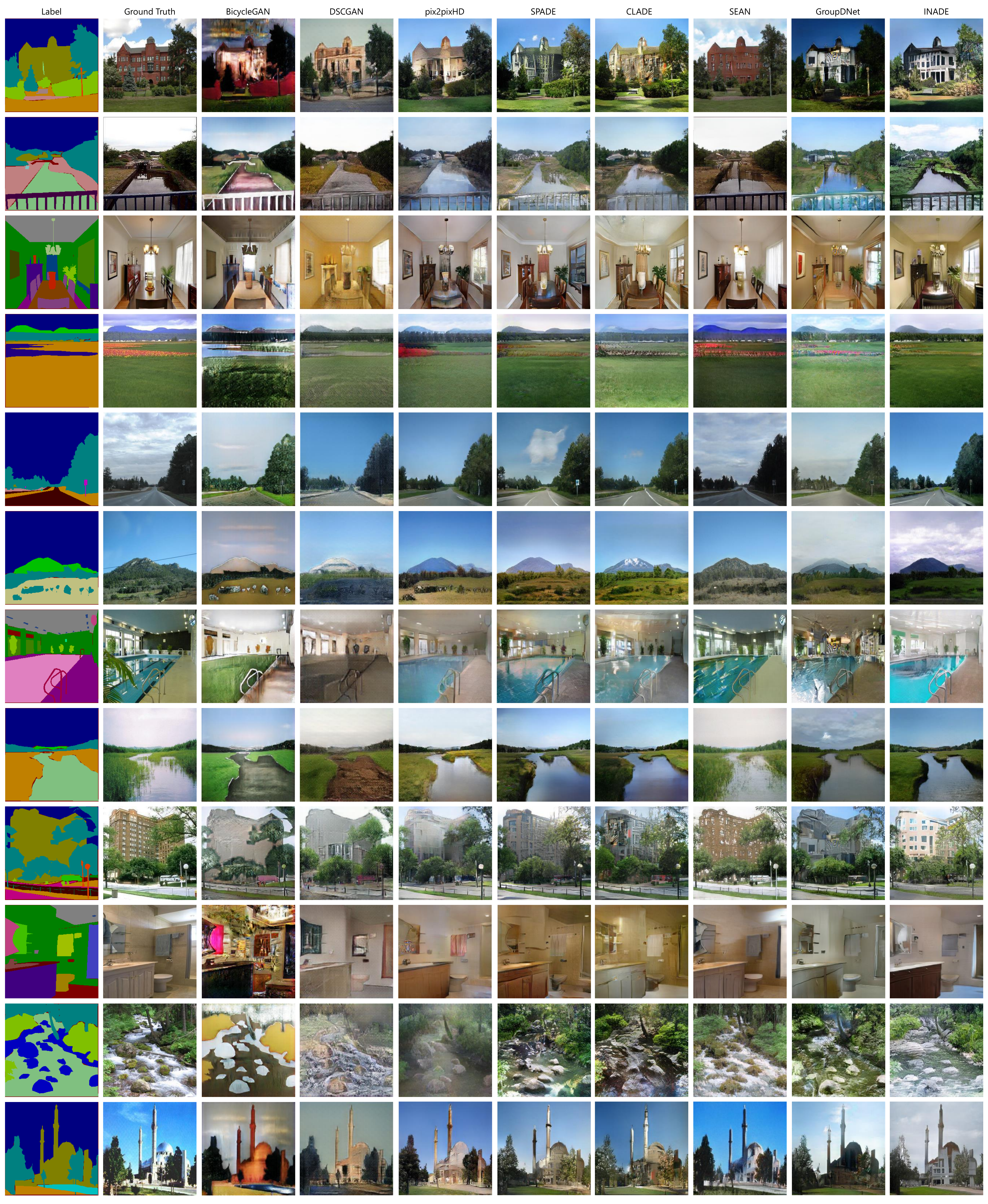}
  \caption{Qualitative comparison of our INADE with previous state-of-the-art methods on the ADE20K dataset.}
  \label{fig:sup_ade}
\end{figure*}

\begin{figure*}[tp]
  \centering
  \includegraphics[width=\linewidth]{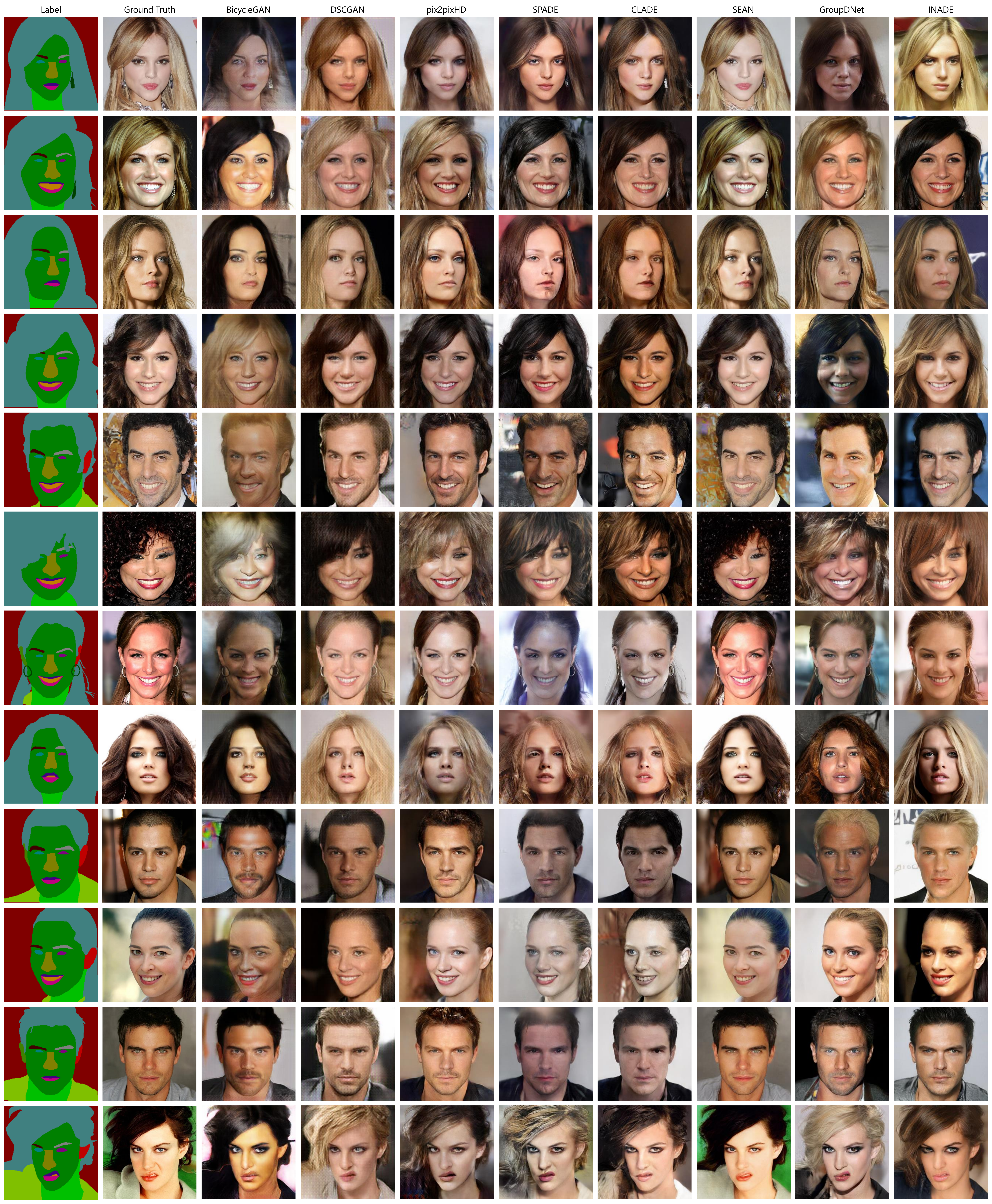}
  \caption{Qualitative comparison of our INADE with previous state-of-the-art methods on the CelebAMask-HQ dataset.}
  \label{fig:sup_celeba}
\end{figure*}

In Figure~\ref{fig:sup_fashion}, Figure~\ref{fig:sup_city}, Figure~\ref{fig:sup_ade}, Figure~\ref{fig:sup_celeba}, we further show additional qualitative comparison results between the proposed INADE and other methods on the \textit{DeepFashion}, \textit{Cityscapes}, \textit{ADE20K} and \textit{CelebAMask-HQ} datasets. These results show that the images quality of INADE is better than or at least comparable to existing methods.

\begin{figure}[tp]
  \centering
  \includegraphics[width=\linewidth]{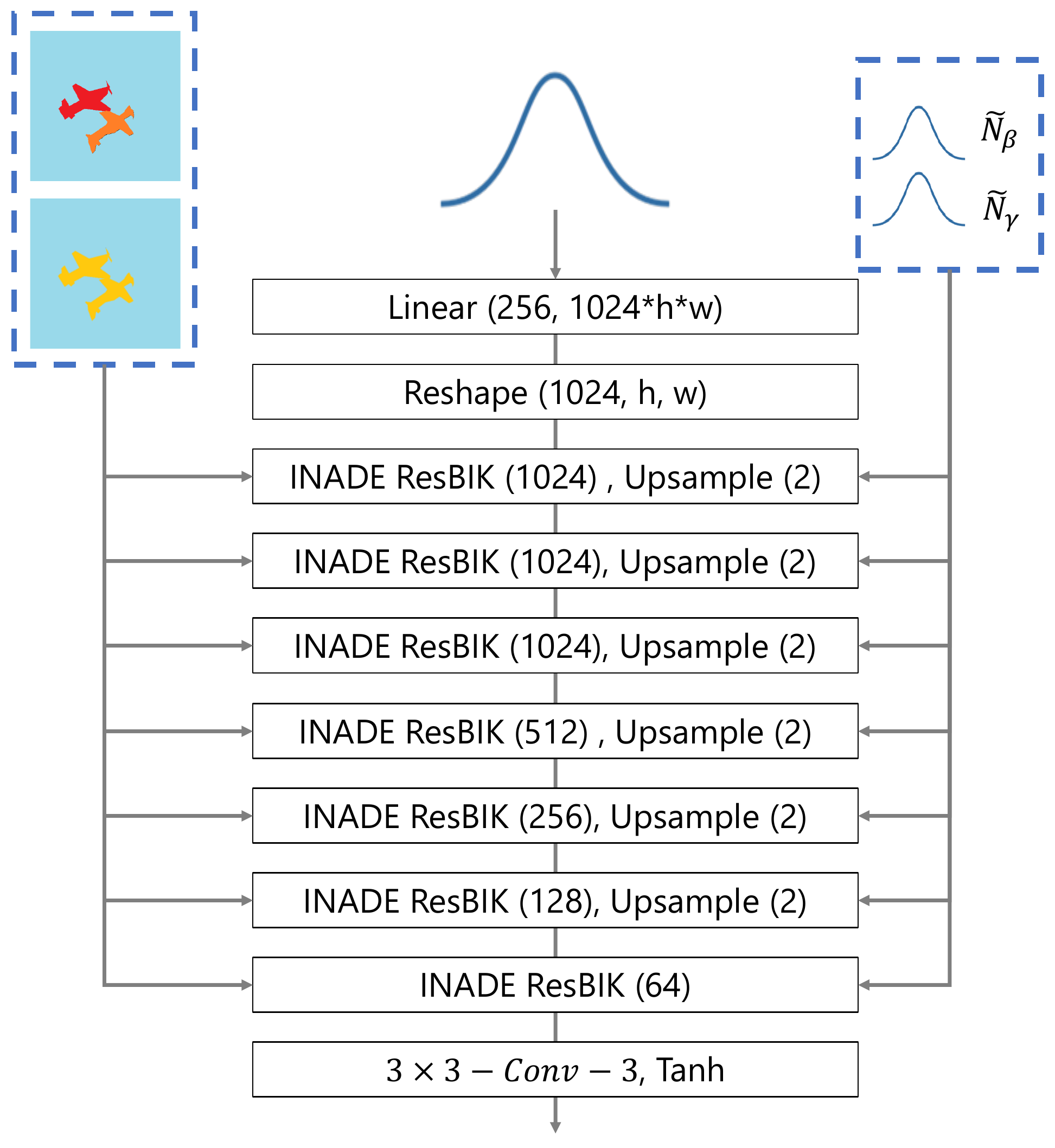}
  \caption{Architecture of our generator network. It consists of a linear transform layer, six INADE ResBIKs with upsampling and a final classification convolution layer. The upsampling operation on the second INADE ResBIK is removed if the resolution of generated images is $256\times 512$. The initial noise ($\tilde{\bm{N}}_\gamma, \tilde{\bm{N}}_\beta$) will be translated through a linear transformation mapping before fed to INADE ResBIKs.}
  \label{fig:generator}
\end{figure}

\begin{table*}[t]
\setlength{\tabcolsep}{1.8mm}
    \centering
    \caption{Comparison of INADE with different $C^o$ on the Cityscapes and CelebAMask-HQ daasets. $\mathbf{P}$, $\mathbf{F}$ and $\mathbf{T}$ represent the generator parameters, FLOPs and run time respectively.}
    %\small
    \footnotesize
    \begin{tabular}{c|c|c|c|c|c|c|c|c|c|c|c|c|c|c}
    \hline
        \multirow{2}{*}{Methods} & \multicolumn{7}{c|}{Cityscapes} & \multicolumn{7}{c}{CelebAMask-HQ}\\
        \cline{2-15}
         & mIoU & acc & FID & LPIPS & $\mathbf{P}$ (M) & $\mathbf{F}$ (G) & $\mathbf{T}$ (s) & mIoU & acc & FID & LPIPS & $\mathbf{P}$ (M) & $\mathbf{F}$ (G) & $\mathbf{T}$ (s)\\
        \hline
        INADE-64 (default) & \textbf{61.02} & \textbf{93.16} & \textbf{38.04} & 0.248 & 77.39 & 75.26 & 0.0486 & \textbf{74.08} & \textbf{94.31} & \textbf{22.55} & 0.365 & 85.12 & 42.18 & 0.0298 \\
        \hline
        INADE-8 & 60.25 & 93.07 & 38.68 & 0.220 & \textbf{76.78} & \textbf{75.23} & \textbf{0.0482} & 73.26 & \textbf{94.31} & 24.58 & 0.350 & \textbf{84.50} & \textbf{42.16} & \textbf{0.0295}\\
        \hline
        INADE-128 & 59.57 & 92.68 & 39.30 & \textbf{0.315} & 78.10 & 75.28 & 0.0497 & 73.48 & 94.28 & 24.88 & \textbf{0.366} & 85.82 & 42.20 & 0.0306\\
        \hline
    \end{tabular}
    \label{tab:ablation}
\end{table*}{}

\end{document}